\documentclass[final]{cvpr}

\usepackage{times}
\usepackage{epsfig}
\usepackage{graphicx}
\usepackage{amsmath}
\usepackage{amssymb}
\usepackage{multicol}

\usepackage{url}

\usepackage[utf8]{inputenc} %
\usepackage[T1]{fontenc}    %
\usepackage{booktabs}       %
\usepackage{amsfonts}       %
\usepackage{nicefrac}       %
\usepackage{microtype}      %
\usepackage{mleftright,xparse}
\usepackage{enumitem}
\usepackage{footnote}
\makesavenoteenv{tabular}
\usepackage{fp}
\usepackage{xr}

\usepackage{color,xcolor}
\usepackage{epsfig}
\usepackage{graphicx}

\usepackage{adjustbox}
\usepackage{array}
\usepackage{booktabs}
\usepackage{colortbl}
\usepackage{float,wrapfig}
\usepackage{hhline}
\usepackage{array,multirow}
\usepackage{subcaption} %

\usepackage{amsmath,amsfonts,amsthm,amssymb}

\usepackage{bm}
\usepackage{nicefrac}
\usepackage{microtype}
\usepackage{dsfont}
\usepackage{amsmath}

\usepackage{changepage}
\usepackage{extramarks}
\usepackage{fancyhdr}
\usepackage{lastpage}
\usepackage{setspace}
\usepackage{soul}
\usepackage{xspace}
\usepackage{makecell}
\usepackage{mfirstuc}
\newtheorem{theorem}{Theorem}

\usepackage{algorithm}
\usepackage{algpseudocode}
\usepackage{capt-of}
\usepackage{enumerate}

\DeclareCaptionFont{ninept}{\fontsize{9pt}{11pt}\selectfont #1}
\usepackage[title]{appendix}
\usepackage{breqn}
\usepackage{amsmath,amsfonts,bm}

\def\eqref#1{equation~\ref{#1}}
\def\1{\bm{1}}

\def\rz{{\textnormal{z}}}

\def\rvh{{\mathbf{h}}}
\def\rvu{{\mathbf{i}}}

\def\rvu{{\mathbf{u}}}

\def\rvx{{\mathbf{x}}}
\def\rvy{{\mathbf{y}}}
\def\rvz{{\mathbf{z}}}

\def\vtheta{{\bm{\theta}}}

\def\vh{{\bm{h}}}

\def\vv{{\bm{v}}}
\def\vw{{\bm{w}}}
\def\vx{{\bm{x}}}
\def\vy{{\bm{y}}}

\def\mI{{\bm{I}}}
\def\mJ{{\bm{J}}}

\def\mW{{\bm{W}}}

\DeclareMathAlphabet{\mathsfit}{\encodingdefault}{\sfdefault}{m}{sl}
\SetMathAlphabet{\mathsfit}{bold}{\encodingdefault}{\sfdefault}{bx}{n}

\def\gE{{\mathcal{E}}}

\def\gG{{\mathcal{G}}}

\def\gV{{\mathcal{V}}}

\def\sA{{A}}

\def\sR{{\mathbb{R}}}

\newcommand{\pdata}{p_{\rm{data}}}
\newcommand{\pmodel}{p_{\rm{model}}}

\newcommand{\E}{\mathbb{E}}

\newcommand{\mhide}[1]{#1}

\newtheorem{lemma}[theorem]{Lemma}

\usepackage{thmtools} % 
\usepackage{thm-restate} % to repeat the same theorem with same number
\usepackage{stfloats}

\usepackage[pagebackref=true,breaklinks=true,colorlinks,bookmarks=false]{hyperref}

\def\cvprPaperID{5597} % *** Enter the CVPR Paper ID here
\def\confYear{CVPR 2021}
\let\svmaketitle\maketitle
\def\maketitle{\svmaketitle\thispagestyle{empty}}

\begin{document}

%%%%%%%%% TITLE
\title{Partition-Guided GANs}
% \author{Mohammadreza\\
% Institution1\\
% Institution1 address\\
% {\tt\small firstauthor@i1.org}
% % For a paper whose authors are all at the same institution,
% % omit the following lines up until the closing ``}''.
% % Additional authors and addresses can be added with ``\and'',
% % just like the second author.
% % To save space, use either the email address or home page, not both
% \and
% Second Author\\
% Institution2\\
% First line of institution2 address\\
% {\tt\small secondauthor@i2.org}
% }
\author{Mohammadreza Armandpour\thanks{Authors contributed equally.}~\textsuperscript{\rm ~1}, Ali Sadeghian$^*$\textsuperscript{\rm 2}, Chunyuan Li \textsuperscript{\rm 3}, Mingyuan Zhou~\textsuperscript{\rm 4} \\
\textsuperscript{\rm 1}Texas A\&M University \qquad \textsuperscript{\rm 2}University of Florida \qquad \\ \textsuperscript{\rm 3}Microsoft Research \qquad \textsuperscript{\rm 4}The University of Texas at Austin\\
{\tt\small armand@stat.tamu.edu, asadeghian@ufl.edu, chunyl@microsoft.com, mzhou@utexas.edu 
%mingyuan.zhou@mccombs.utexas.edu
}
% For a paper whose authors are all at the same institution,
% omit the following lines up until the closing ``}''.
% Additional authors and addresses can be added with ``\and'',
% just like the second author.
% To save space, use either the email address or home page, not both
}

\maketitle
% \pagenumbering{gobble}
% \pagestyle{empty}
% \thispagestyle{empty}
%%%%%%%%% ABSTRACT
\begin{abstract}
Despite the success of Generative Adversarial Networks (GANs), their training suffers from several well-known problems, including mode collapse and difficulties learning a disconnected set of manifolds. In this paper, we break down learning complex high dimensional distributions to simpler sub-tasks, supporting diverse data samples. Our solution relies on designing a partitioner that breaks the space into smaller regions, each having a simpler distribution, and training a different generator for each partition. This is done in an unsupervised manner without requiring any labels. 

We formulate two desired criteria for the space partitioner that aid the training of our mixture of generators: 1)~to produce connected partitions and 2) provide a proxy of distance between partitions and data samples, along with a direction for reducing that distance. These criteria are developed to avoid producing samples from places with non-existent data density, and also facilitate training by providing additional direction to the generators. We develop theoretical constraints for a space partitioner to satisfy the above criteria. Guided by our theoretical analysis, we design an effective neural architecture for the space partitioner that empirically assures these conditions. Experimental results on various standard benchmarks show that the proposed unsupervised model outperforms several recent methods. 
\end{abstract}

\section{Introduction}

Generative adversarial networks (GANs)~\cite{goodfellow2014generative} have gained remarkable success in learning the underlying distribution of observed samples. However, their training is still unstable and challenging, especially when the data distribution of interest is multimodal. This is particularly important due to both empirical and theoretical evidence that suggests real data also conforms to such distributions~\cite{narayanan2010sample, tenenbaum2000global}. 

Improving the vanilla GAN, both in terms of training stability and generating high fidelity images, has been the subject of great interest in the machine learning literature~\cite{arjovsky2017wasserstein,denton2015deep,gulrajani2017improved,liu2019normalized,mao2017least,mescheder2018training, mirza2014conditional,radford2015unsupervised,sadeghian2019sophie,salimans2016improved, xiao2018bourgan}. One of the main problems is \textit{mode collapse}, where the generator fails to capture the full diversity of the data. Another problem, which hasn't been fully explored, is the \textit{mode connecting} problem~\cite{khayatkhoei2018disconnected, tanielian2020learning}. As we explain in detail in Section~\ref{sec:gan_problems}, this phenomenon occurs when the GAN generates samples from parts of the space where the true data is non-existent, caused by using a continuous generator to approximate a distribution with disconnected support. Moreover, GANs are also known to be hard to train due to the unreliable gradient provided by~the~discriminator. 

\begin{figure}
\begin{subfigure}{.235\textwidth}
  \centering
  \caption*{~~~~Real~~~~~~~~~~~~~Generated} \vspace{-6pt}
  \includegraphics[width=.48\linewidth]{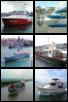}
  \includegraphics[width=.48\linewidth]{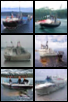}
  \caption{Partition 13}
  \label{fig:sfig1}
\end{subfigure}\hfill
\begin{subfigure}{.235\textwidth}
  \centering
  \caption*{~~~~Real~~~~~~~~~~~~~Generated} \vspace{-6pt}
  \includegraphics[width=.48\linewidth]{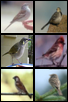}
  \includegraphics[width=.48\linewidth]{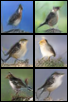}
  \caption{Partition 47}
  \label{fig:sfig2}
\end{subfigure}
\begin{subfigure}{.235\textwidth}
  \centering
  \caption*{~~~~Real~~~~~~~~~~~~~Generated} \vspace{-6pt}
  \includegraphics[width=.48\linewidth]{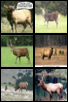}
  \includegraphics[width=.48\linewidth]{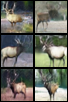}
  \caption{Partition 69}
  \label{fig:sfig3}
\end{subfigure}\hfill
\begin{subfigure}{.235\textwidth}
  \centering
  \caption*{~~~~Real~~~~~~~~~~~~~Generated} \vspace{-6pt}
  \includegraphics[width=.48\linewidth]{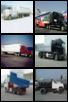}
  \includegraphics[width=.48\linewidth]{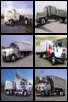}
  \caption{Partition 85}
  \label{fig:sfig4}
\end{subfigure}%
\caption{Examples of unsupervised partitioning and their corresponding real/generated samples on the CIFAR-10 dataset.}
\label{fig:cifar_intro}
\end{figure}

Our solution to alleviate the aforementioned problems is introducing an unsupervised space partitioner and training a different generator for each partition. Figure~\ref{fig:cifar_intro} illustrates real and generated samples from several inferred partitions. 

Having multiple generators, which are focused on different parts/modes of the distribution, reduces the chances of missing a mode. This also mitigates mode connecting because the mixture of generators is no longer restricted to be a continuous function responsible for generating from a data distribution with potentially disconnected manifolds. In this context, an effective space partitioner should place disconnected data manifolds in different partitions. Therefore, assuming semantically similar images are in the same connected manifold, we use contrastive learning methods to learn semantic representations of images and partition the space using these embeddings.

We show that the space partitioner can be utilized to define a distance between points in the data space and partitions. The gradient of this distance can be used to encourage each generator to focus on its corresponding region by providing a direction to guide it there. In other words, by penalizing a generator when its samples are far from its partition, the space partitioner can \textit{guide} the generator to its designated region. Our partitioner's guide is particularly useful where the discriminator does not provide a reliable gradient, as it can steer the generator in the right direction.

However, for a reliable \textit{guide}, the distance function must follow certain characteristics, which are challenging to achieve. For example, to avoid misleading the GANs' training, the distance should have no local optima outside the partition. In Section~\ref{sec:partition_guide_gan}, we formulate sufficient theoretical conditions for a desirable metric and attain them by enforcing constraints on the architecture of the space partitioner. This also guarantees connected partitions in the data space, which further mitigates mode connecting as a by-product. 

\vspace{3pt}

We perform comprehensive experiments on StackedMNIST~\cite{lin2018pacgan,liu2020diverse,srivastava2017veegan}, CIFAR-10~\cite{krizhevsky2009learning}, STL-10~\cite{coates2011stl10} and ImageNet~\cite{russakovsky2015imagenet} without revealing the class labels to our model. We show that our method, \textit{Partition-Guided Mixture of GAN} (PGMGAN), successfully recovers all the modes and achieves higher Inception Score (IS) \cite{salimans2016improved} and Frechet Inception Distance (FID) \cite{heusel2017gans} than a wide range of supervised and unsupervised methods. 

Our contributions can be summarized as: 
\begin{itemize}
    \item Providing a theoretical lower bound on the total variational distance of the true and estimated densities % and estimated one %density
    using a single generator.
    
    \item Introducing a novel differentiable space partitioner and demonstrating that simply training a mixture of generators on its partitions alleviates mode collapse/connecting.
    
    \item Providing a practical way (with theoretical guarantees) to guide each generator to produce samples from its designated region, further improving mode collapse/connecting. Our experiments show significant improvement over relevant baselines in terms of FID and IS, confirming the efficacy of our model. 
    
    \item Elaborating on the design of our loss and architecture by making connection to supervised GANs that employ a classifier. We explain how PGMGAN avoids their~shortcomings.
\end{itemize}

\section{Mode connecting problem}
\label{sec:gan_problems}

Suppose the data distribution is supported on a set of disconnected manifolds embedded within a higher-dimensional space. Since continuous functions preserve the space connectivity~\cite{kelley2017general}, one can never expect to have an exact approximation of this distribution by applying a continuous function ($G_{\vtheta}$) to a random variable with a connected support. Furthermore, if we restrict $G_{\vtheta}$ to the class of c-Lipschitz functions, the distance between the true density and approximated will always remain more than a certain positive value. In fact, the generator would either have to discard some of the data manifolds or connect the manifolds. The former can be considered a form of \textit{mode collapse}, and we refer to the latter as the~\textit{mode~connecting}~problem. 

The following theorem formally describes the above statement and provides a lower bound for the total variation distance between the true and estimated densities.

\begin{restatable}[]{thm}{ModeConnectingThm}
\label{thm:mode_connecting}
Let $\pdata$ be a distribution supported on a set of disjoint manifolds $\mathcal{M}_1, \dots, \mathcal{M}_k$ in $\sR^d$, and $[\pi_1, \dots, \pi_k]$ be the probabilities of being from each manifold. Let $G_{\vtheta}$ be a c-Lipschitz function, and $\pmodel$ be the distribution of $G_{\vtheta}(\rvz)$, where $\rvz \sim \mathcal{N}(0, \mI_n)$, then:
$$d_{TV}(\pdata, \pmodel)\geq \sum |\pi_i -p_i| \geq \delta$$

\noindent where $d_{TV}$ is the total variation distance and:
\begin{align*}
    \pi_i^* &:= \min(\pi_i, 1-\pi_i)\\
    p_i &:= \pmodel(\mathcal{M}_i)\\
    \delta \, &:= \max_i{ \{ \pi_i^* - \Phi( \Phi^{-1}(\pi_i^*) - d_i/c) \} }\\
    d_i &:= \inf\{||\vx-\vy ||~~ | ~\vx \in \mathcal{M}_i, \vy \in \mathcal{M}_j, j\neq i \}
\end{align*}
$d_i$ is the distance of manifold $\mathcal{M}_i$ from the rest, and $\Phi$ is the CDF of the univariate standard normal distribution. Note $\delta$ is strictly larger than zero iff $ \exists i: \, d_i,\pi_i^* \neq 0$. 
\end{restatable}

According to Theorem~\ref{thm:mode_connecting}, the distance between the estimated density and the data distribution can not converge to zero when $G_{\vtheta}$ is a Lipschitz function. It is worth noting that this assumption holds in practice for most neural architectures as they are a composition of simple Lipschitz functions. Furthermore, most of the state-of-the-art GAN architectures (e.g., BigGAN~\cite{brock2018large} or SAGAN~\cite{zhang2019self}) use spectral normalization in their generator to stabilize their training, which promotes Lipschitzness.

\section{Related work}
% TODO Add the papers used as STL benchmarks
% TODO Add: (refrences added and cited but we did not explain)
% [2] Xiao, Chang, Peilin Zhong, and Changxi Zheng. "BourGAN: Generative networks with metric embeddings." NeurIPS 2018.
% [3] Liu, Shaohui, Xiao Zhang, Jianqiao Wangni, and Jianbo Shi. "Normalized diversification." CVPR 2019.
% They try to generate images such that its distribution matches the distribution of the pairwise distance of the data samples. We conducted experiments on CIFAR10 and CelebA with state-of-the-art methods via the off-the-shelf library1 . Because there exist relatively dense samples in this setting, conventional methods fit the problem relatively well.

Apart from their application in computer vision~\cite{hoffman2018cycada,huang2018multimodal,isola2017image, karras2018progressive,karras2019style,wang2016generative,zhang2017stackgan,zhu2017unpaired}, GANs have also been been employed in natural language processing~\cite{ li2017adversarial, lin2017adversarial, yu2017seqgan}, medicine~\cite{schlegl2017unsupervised, killoran2017generating} and several other fields~\cite{florensa2018automatic,marouf2020realistic,pascual2017segan}. Many of recent research have accordingly focused on providing ways to avoid the problems discussed in Section~\ref{sec:gan_problems} \cite{lin2018pacgan, metz2017unrolled}.

\textbf{Mode collapse}
% unroll gan
For instance, Metz et al.~\cite{metz2017unrolled}~unroll the optimization of the discriminator to obtain a better estimate of the optimal discriminator at each step, which remedies mode collapse. However, due to high computational complexity, it is not scalable to large datasets. VEEGAN~\cite{srivastava2017veegan} adds a reconstruction term to bi-directional GANs~\cite{dumoulin2016adversarially, donahue2016adversarial} objective, which does not depend on the discriminator. This term can provide a training signal to the generator even when the discriminator does not. PacGAN~\cite{lin2018pacgan}~changes the discriminator to make decisions based on a pack of samples. This change mitigates mode collapse by making it easier for the discriminator to detect lack of diversity and naturally %penalizing 
penalize
the generator when mode collapse happens. Lucic et al.~\cite{lucic2019high}, motivated by the better performance of supervised-GANs, propose using a small set of labels and a semi-supervised method to infer the labels for the entire data. They further improve the performance by utilizing an auxiliary rotation loss similar to that of RotNet~\cite{gidaris2018unsupervised}.

\textbf{Mode connecting} Based on Theorem~\ref{thm:mode_connecting}, to avoid mode connecting one has to either use a latent variable $\rvz$ with a disconnected support, or allow $G_{\vtheta}$ to be a discontinuous function~\cite{hoang2018mgan, khayatkhoei2018disconnected, kundu2019gan,  liu2020diverse, sage2018logo}. 

To obtain a disconnected latent space, DeLiGAN~\cite{gurumurthy2017deligan} samples $\rvz$ from a mixture of Gaussian, while  Odena et al.~\cite{odena2017conditional} add a discreet dimension to the latent variable. Other methods dissect the latent space post-training using some variant of rejection sampling, for example, Azadi et al.~\cite{azadi2019discriminator} perform rejection sampling based on the discriminator's score, and Tanielian et al.~\cite{tanielian2020learning} reject the samples where the generator's Jacobian is higher than a certain threshold. 

The discontinuous generator method is mostly achieved by learning multiple generators, with the primary motivation being to remedy mode-collapse, which also reduces mode connecting. Both MGAN~\cite{hoang2018mgan} and DMWGAN~\cite{khayatkhoei2018disconnected} employ $K$ different generators while penalizing them from overlapping with each other. %if they overlap. 
However, these works do not explicitly address the issue when some of the data modes are not being captured. Also, as shown in Liu et al.~\cite{liu2020diverse}, MGAN is quite sensitive to the choice of $K$. By contrast, Self-Conditioned GAN~\cite{liu2020diverse} clusters the space using the discriminator's final layer and uses the labels as self-supervised conditions. However, in practice, their clustering does not seem to be reliable (e.g., in terms of NMI for labeled datasets), and the features highly depend on the choice of the discriminator's architecture. In addition, there is no guarantee that the generators will be guided to generate from their assigned clusters. GAN-Tree~\cite{kundu2019gan} uses hierarchical clustering to address continuous multi-modal data, with the number of parameters increasing linearly with the number of clusters. Thus it is limited to very few cluster numbers (e.g., 5) and can only capture a~few~modes. 

Another recently expanding direction explores the benefit of using image augmentation techniques for generative modeling. Some works simply augment the data using various perturbations (e.g., random crop, horizontal flipping)~\cite{karras2020analyzing}. Others~\cite{chen2019self, lucic2019high, zhao2020image} incorporated regularization on top of the augmentations, for example CRGAN~\cite{zhang2020consistency} enforces consistency for different image perturbations. ADA~\cite{karras2020training} processes each image using non-leaking augmentations and adaptively tunes the augmentation strength while training. These works are orthogonal to ours and can be combined with our method.  

% Furthermore, the simplicity of our method allows it to be easily combined with a variety of these techniques

\section{Method}
\label{sec:method}

This section first describes how GANs are trained on a partitioned space using a mixture of generators/discriminators and the unified objective function required for this goal. We then explain our differentiable space partitioner and how we guide the generators towards the right region. We conclude the section by making connections to supervised GANs, which use an auxiliary classifier~\cite{miyato2018cgans, odena2017conditional}. \\

\textbf{Multi-generator/discriminator objective:} Given a partitioning of the space, we train a generator ($G_i$) and a discriminator ($D_i$) for each region. To avoid over-parameterization and allow information sharing across different regions, we employ parameter sharing across different $G_i$ ($D_i$) by tying their parameters except the input (last) layer.  The mixture of these generators serves as our main generator~$G$. We use the following objective function to train our GANs:
\begin{equation}
\label{eq:maineq}
\sum_i^k \pi_{i} \left[ \min_{G_i} \max_{D_i}  V(D_i, G_i, \sA_i) \right]
\end{equation}
 where $\sA_1, \sA_2,..., \sA_k$ is a partitioning of the space, $\pi_{i} := p_{data}(\rvx \in \sA_i)$  and:

\begin{align}
    V(D, G, \sA) = \, & \mathbb{E}_{\bm{x} \sim p_{\text{data}}(\bm{x} | \vx \in \sA )}[\log D(\bm{x})] \,\, + \nonumber \\
    & \mathbb{E}_{\bm{z} \sim p_{\bm{z}}(\bm{z} | G(\bm{z}) \in \sA)}[\log (1 - D(G(\bm{z})))]
\end{align}
We motivate this objective by making connection to the Jensen--Shannon distance (${\rm JSD}$) between the distribution of our mixture of generators and the data distribution in the following Theorem.
% The details of this theorem and its proof are provided the~Appendix.
\begin{restatable}[]{thm}{MixtureThm}
\label{thm:mixture}
Let $P = \sum_i^k \pi_i p_i$ , $Q = \sum_i^k \pi_i q_i$ , and $\sA_1, \sA_2,..., \sA_K$ be a partitioning of the space, such that the support of each distribution $p_i$ and $q_i$ is $\sA_i$. Then:\\
\begin{equation}
\label{eq:theorem1}
{\rm JSD}(P \parallel Q) = \sum_i \pi_i {\rm JSD}(p_i \parallel q_i)
\end{equation}
\end{restatable}

\subsection{Partition GAN}
\label{sec:parition_gan}

\textbf{Space Partitioner:} Based on Theorem~\ref{thm:mode_connecting}, an ideal space partitioner should place disjoint data manifolds in different partitions to avoid mode connecting (and consequently mode collapse). It is also reasonable to assume that semantically similar data points lay on the same manifold. Hence, we train our space partitioner using semantic embeddings. 

We achieve this goal in two steps: 1) Learning an unsupervised representation for each data point, which is invariant to transformations that do not change the semantic meaning. 
2) Training a partitioner based on these features, where data points with similar embeddings are placed in the~same partition. \\

% \vspace{3pt}
\textit{Learning representations:} We follow the self-supervised literature~\cite{chen2020simple, chen2020big,he2020momentum} to construct image representations.

These methods generally train the networks via maximizing the agreement between augmented views (e.g., random crop, color distortion, rotation, etc.) of the same scene, while minimizing the agreement of views from different scenes. To that end, they optimize the following contrastive loss:
\begin{equation}
\label{eq:contrastive_loss}
\sum_{(i,j) \in P} \log \frac{\exp(\mathrm{sim}(\rvh_i, \rvh_j)/\tau)}{\sum_{k=1}^{2N} \1_{k \neq i}\exp(\mathrm{sim}(\rvh_i, \rvh_k)/\tau)}
\end{equation}

\noindent where $\rvh$ is the embedding for image $\rvx$, $(i, j)$ is a positive pair (i.e., two views of the same image) and $(i, k)$ refers to negative pairs related to two different images. We refer to this network as pretext, implying the task being solved is not of real interest but is solved only for the true purpose of learning a suitable data representation.\\

% \vspace{3pt}
\textit{Learning partitions:} To perform the partitioning step, one can directly apply K-means on these semantic representations. However, this may result in degenerated clusters where one partition contains most of the data \cite{caron2018deep, wvangansbeke2020scan}. 
Inspired by Van Gansbek et al.~\cite{wvangansbeke2020scan}, to mitigate this challenge, we first make a k-nearest neighbor graph $\gG =(\gV, \gE)$ based on the representations $\rvh$ of the data points.
We then train an unsupervised model that motivates connected points to reside in the same cluster and  disconnected points to reside in distinct clusters. More specifically, we train a space partitioner $S:\sR^d \to [0, 1]^k$
to maximize:
\begin{align}
\label{eq:cluster}
        \sum_{(i, j) \in \gE} & \log \left(S(\rvx_i)^{T} \cdot S(\rvx_j)\right) - \nonumber \\
    & \alpha \sum_{i \in \gV} H_{C}(S(\rvx_i))) + \beta  H_{C}(\sum_{i \in \gV} \frac{S(\rvx_i)}{N} ) 
\end{align}
where $H_C(.)$ is the entropy function of the categorical distribution based on its probability vector. The first term in Equation~\ref{eq:cluster} motivates the neighbors in $\gG$ to have similar class probability vectors, with the $\log$ function used to significantly penalize the classifier if it assigns dissimilar probability vector to two neighboring points. The last term is designed to avoid placing all the data points in the same category by motivating the average cluster probability vector to be similar to the uniform distribution. 
The middle term is intended to promote the probability vector for each data point to significantly favor one class over the other. This way, we can be more confident about the cluster id of each data point. Furthermore, if the average probability of classes has a homogeneous mean (because of the last term), we can expect the number of data points in each class to~not~degenerate.

To train $S$ efficiently, both in term of accuracy and computational complexity, we initialize $S$ using the already trained network of unsupervised features. More specifically, for:
$$\rvh = W_2^{\textit{pretext}}\sigma(W_1^{\textit{pretext}}\phi_{0}(\rvx))$$
we initialize $S$ as follows:
$$S^{\textit{init}}(\rvx) = \textit{softmax}( W_0^{\textit{partitioner}}\phi_{0}(\rvx))$$
where $\sigma$ is an activation function, and $W_0^{\textit{partitioner}}$ is a randomly initialized matrix; we ignore the bias term here for %the sake of
brevity. We drop the sub index $0$, from  $W_0^{\textit{partitioner}}$ and $ \phi_{0}$ to refer to their post-training versions. Given a fully trained $S$, each point $\rvx$ is assigned to partition $A_i$, based on the argmax of the probability vector of $S(\rvx)$.

\subsection{Partition-Guided GAN}
\label{sec:partition_guide_gan}

In this section we describe the design of \textit{guide} and its properties. As stated previously, we want to guide each generator $G_i$ to its designated region $A_i$ by penalizing it the farther its current generated samples are from $A_i$. 

A simple, yet effective proxy of measuring this distance can be attained using the already trained space partitioner. Let $f_i$s denote the partitioner's last layer logits, expressed as
$$[f_1(\rvx), \dots, f_k(\rvx)]^T := W^{\textit{partitioner}}\phi(\rvx).$$

\noindent and define the desired distance as:
\begin{equation}\label{eq:relu}
    R_i(\rvx) := \sum_c (f_c(\rvx) - f_i(\rvx))_{+} 
\end{equation} 

\textit{Property $1$.} It is easy to show that for any generated sample $\rvx$, $R_i(\rvx)$ achieves a larger value, the less likely $S$ believes $\rvx$ to be from partition $A_i$. This is clear from how we defined $R_i$, the more probability mass $S(\rvx)$ assigns any class $c \neq i$, the larger the value of $R_i(\rvx)$. 

\textit{Property $2$.} It is also straightforward to see that $R_i(\rvx)$ is always non-negative and obtains its minimum (zero) only on the $A_i$\textsuperscript{th} partition:
\begin{align}
    \begin{split}
        \rvx \in \sA_i \iff &  f_i(\rvx) \geq f_c(\rvx); \quad \forall c \in [1:k] \\
        \iff &  R_i(\rvx) = \sum_c (f_c(\rvx) - f_i(\rvx))_{+} = 0
    \end{split}
\end{align}

Therefore, we guide each generator $G_i$ to produce samples from its region by adding a penalization term to its canonical objective function:
\begin{align}
\label{eq:regguide}
       \min_{G_i} \sum_{j=1}^n 
              \log (1-D_i  &(G_i  (\bm{z}^{(j)})))) \nonumber \\ 
    + \lambda &\sum_j R_i(G_i(\bm{z}^{(j)}))/n . 
\end{align}

Intuitionally, $G_i$ needs to move its samples towards partition $A_i$ in order to minimize the newly added term. Fortunately, given the differentiability of $R_i(.)$ with respect to its inputs and \textit{Property $1$}, $R_i$ can provide the direction for $G_i$ to achieve that goal. 

It is also worth noting that $R_i$ should not interfere with the generator/discriminator as long as $G_i$'s samples are within $\sA_i$. Otherwise, this may lead to the second term favoring parts of $\sA_i$ over others and conflicting with the discriminator. \textit{Property $2$} assures that learning the distribution of $p_{data}$ over $\sA_i$ remains the responsibility of $D_i$. We also use this metric to ensure each trained generator $G_i$ only draws samples from within its region by only accepting samples with $R_i(\rvx)$ being equal to zero. 

A critical point left to consider is the possibility of $G_i$ getting fooled to generate samples from outside $\sA_i$, by falling in local optima of $R_i(\rvx)$. In the remaining part of this section, we will explain how the architecture design of the space partitioner $S$ avoids this issue. In addition, it will also guarantee the norm of the gradient provided by $R_i$ to always be above a certain threshold. \\

\textbf{Avoiding local optima:} We can easily obtain a guide $R_i$ with no local optima, if achieving a good performance for the partitioner was not important. For instance, a simple single-linear-layer neural network as $S$ would do the trick.
The main challenge comes from the fact that we need to perform well on partitioning~\footnote{Accurately put different manifolds in different partitions.}, which usually requires deep neural networks while avoiding local optima. We fulfill this goal by first finding a sufficient condition to have no local optima and then trying to enforce that condition by modifying the ResNet~\cite{he2016deep} architecture. 

The following theorem states the sufficient condition:
\begin{restatable}[]{thm}{SuffThm}
\label{thm:suff}
Let $\phi(\rvx): \sR^d \to \sR^d$ be a $C^{1}$ (differentiable with continuous derivative) function, $W^{\textit{partitioner}} \in \sR^{k \times d}$, and $R_i$ as defined in Eq~\ref{eq:relu}. If there exists $c_0>0$, such that:
$$ \forall \, \rvx, \rvy \in \sR^d, ~~~~  c_0 ||\rvx - \rvy|| \leq ||\phi(\rvx) - \phi(\rvy)||,$$
then for every $i \in [1:k]$, every local optima of $R_i$ is a global optima, and there exists a positive constant $b_0 > 0$ such that:
$$\forall \rvx \in \sR^d \setminus A_i, ~~~~ b_0 \leq ||\nabla R_i(\rvx) ||$$
where $A_i= \{\rvx | \rvx \in \sR^d, R_i(\rvx)= 0 \}$. Furthermore $A_i$ is a connected set for all $i$'s.
\end{restatable}
\noindent The proof is provided in the Appendix. Next we describe how to satisfy this constraint in practice.\\

\begin{figure}[t]
\includegraphics[width=0.45\textwidth]{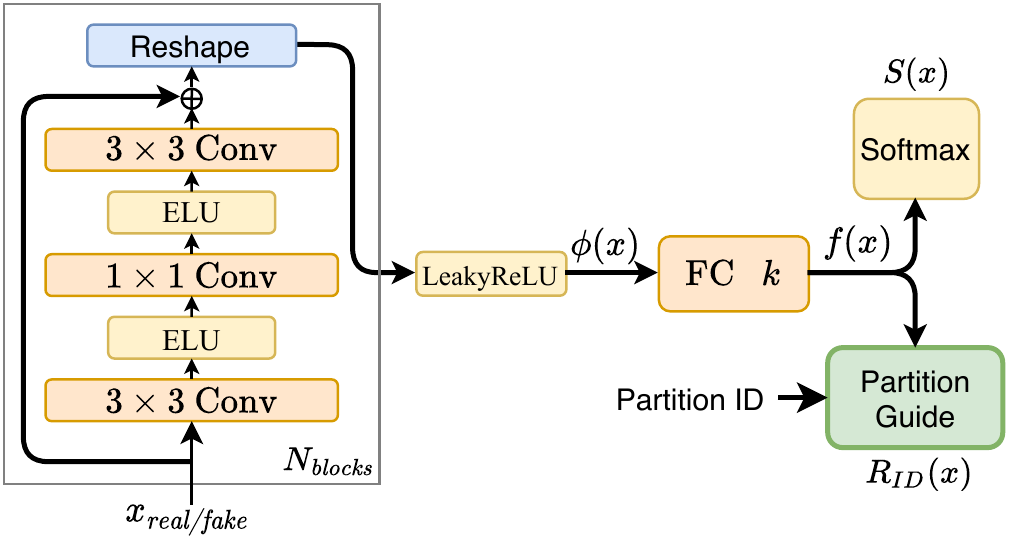}
\caption{Diagram of proposed partitioner and guide. We employ spectral normalization for each convolutional layer to make each layer (as a function) have Lipschitz constant of less than one. The details of our architecture is provided in the Appendix.}
\label{fig:guide_diagram}
\end{figure}

Motivated by the work of Behrmann et al.~\cite{behrmann2019invertible} who design an invertible network without significantly sacrificing their classification accuracy, we implement $\phi$ by stacking several residual blocks,
$\phi(\rvx)= B_T \circ B_{T-1} \circ \cdots \circ B_1(\rvx)$, where:
$$B_{t+1}(\rvx^{(t)}) = \rvx^{(t+1)} := \rvx^{(t)} + \psi_t(\rvx^{(t)})$$ 
and $\rvx^{(t)}$ refers to the out of the $t$\textsuperscript{th} residual block. Figure~\ref{fig:guide_diagram}, gives an overview of the proposed architecture.

We model each $\psi_t$ as a series of $m$ convolutional layers, each having spectral norm $L < 1$ intertwined by $1$-Lipschitz activation functions (e.g., ELU, ReLU, LeakyReLU). Thus it can be easily shown for all $\rvx^{(t)}, \rvy^{(t)} \in \sR^d $:

$$(1- L^m) ||\rvx^{(t)} - \rvy^{(t)}|| \leq ||B_t(\rvx^{(t)}) - B_t(\rvy^{(t)})||.$$
This immediately results in the condition required in Theorem \ref{thm:suff} by letting $c_0:=(1- L^m)$.

\subsection{Connection to supervised GANs}

In this section, we make a connection between our unsupervised GAN and some important work in the supervised regime. This will help provide better insight into why the mentioned properties of guide are important. Auxiliary Classifier GAN~\cite{odena2017conditional} has been one of the well-known supervised GAN methods which uses the following objective function:

\resizebox{0.98\linewidth}{!}{
\begin{minipage}{\linewidth}
\centering
\begin{align*}
   & \underset{{G,C}}{\min}~\underset{{D}}{\max} \; \; \mathcal{L}_{\text{AC}}(G,D,C) = \\
   & \underbrace{\underset{X\sim P_{X}}{\E} [\log D(X)]+\underset{Z\sim P_Z, Y\sim P_Y}{\E} [\log(1- D(G(Z,Y)))]}_{\textcircled{\small a}} \nonumber \\
   & -\lambda_c\underbrace{\underset{(X,Y)\sim P_{XY}}{\E} [\log C(X,Y)]}_{\textcircled{\small b}}-\lambda_c\underbrace{\underset{Z\sim P_{Z},Y\sim P_{Y}}{\E} [\log(C(G(Z,Y),Y))]}_{\textcircled{c}}
    \label{Eq:acgan}
\end{align*}
\end{minipage}
}

It simultaneously learns an auxiliary classifier $C$ as well as $D/G$. Other works have also tried fine-tuning the generator using a pre-trained classifier~\cite{miyato2018cgans}. The term $a$ is related to the typical supervised conditional GAN, term $b$ motivates the classifier to better classify the real data. The term $c$ encourages $G$ to generate images for each class such that the classifier considers them to belong to that class with a~high~probability. %\mz{For part (a), is Y missing from the first term? Does D know whether it is supposed to discriminator images from a particular class?}

The authors motivate adding this term as it can provide further gradient to the generator $G(\cdot|Y)$ to generate samples from the correct region $P_X(\cdot|Y)$. However, recent works~\cite{gong2019twin, shu2017ac} show this tends to motivate $G$ to down-sample data points near the decision boundary of the classifier. It has also been shown to reduce sample diversity and does not behave well when the classes share overlapping regions~\cite{gong2019twin}. 

Our space partitioner acts similar to the classifier in these GANs, with the term $c$ sharing some similarity with our proposed \textit{guide}. In contrast, our novel design of $R_i(\cdot)$ enjoys the benefits of the classifier based methods (providing gradient for the generator) but alleviates its problems. Mainly because 1) It provides gradient to the generator to generate samples from its region. At the same time, due to having no local optima (only global optima), it does not risk the generator getting stuck where it is not supposed to. 2) Within regions, our guide does not mislead the generator to favor some samples over others. 3) Since the space partitioner uses the partition labels as ``class'' id, it does not suffer from the overlapping classes problem, and naturally, it does not require supervised labels. 

We believe our construction of the modified loss can also be applied to the supervised regime to avoid putting the data samples far from the boundary. In addition, combining our method with the supervised one, each label itself can be partitioned into several segments. We leave the investigation of this modification to future research.

\begin{table*}[h]
\centering%

\vspace{-3mm}
\resizebox{%
    1.60\columnwidth
}{!}
{%  
  \begin{tabular}{lrrrrr}
    \toprule
    \multicolumn{1}{c}{} &
    \multicolumn{2}{c}{\textbf{Stacked MNIST}} &
    \multicolumn{3}{c}{\textbf{CIFAR-10}} \\ 
    \cmidrule(r){2-3}
    \cmidrule(r){4-6}
            & \makecell{Modes \\ (Max 1000)}$~\uparrow$ & \multicolumn{1}{c}{Reverse KL$~\downarrow$}  & \multicolumn{1}{c}{FID$~\downarrow$}  & \multicolumn{1}{c}{IS$~\uparrow$}  & \multicolumn{1}{c}{Reverse KL$~\downarrow$}    \\
    \midrule
    
    GAN~\cite{goodfellow2014generative} & 
    $133.4\mhide{\pm 17.70}$ & $2.97 \mhide{\pm 0.216}$ & $28.08 \mhide{\pm 0.47}$ &  ${6.98} \mhide{\pm 0.062}$ & ${0.0150} \mhide{\pm 0.0026}$ \\
    PacGAN2~\cite{lin2018pacgan}  & 
    $\textbf{1000.0} \mhide{\pm 0.00}$ & $0.06 \mhide{\pm 0.003}$ & $27.97 \mhide{\pm 0.63}$ &  ${7.12} \mhide{\pm 0.062}$ & ${0.0126} \mhide{\pm 0.0012}$ \\
    PacGAN3~\cite{lin2018pacgan}  & 
    $\textbf{1000.0} \mhide{\pm 0.00}$ & $0.06 \mhide{\pm 0.003}$ &  $32.55 \mhide{\pm 0.92}$ &  ${6.77} \mhide{\pm 0.064}$ & ${0.0109} \mhide{\pm 0.0011}$ \\
    PacGAN4~\cite{lin2018pacgan} & 
    $\textbf{1000.0} \mhide{\pm 0.00}$ & $0.07 \mhide{\pm 0.005}$ & $34.16 \mhide{\pm 0.94}$ &  ${6.77} \mhide{\pm 0.079}$ & ${0.0150} \mhide{\pm 0.0005}$ \\
    Logo-GAN-AE~\cite{sage2018logo} & 
    $\textbf{1000.0} \mhide{\pm 0.00}$ & $0.09 \mhide{\pm 0.005}$ &  $32.49 \mhide{\pm 1.37}$ &  ${7.05} \mhide{\pm 0.073}$ & ${0.0106} \mhide{\pm 0.0005}$ \\
    % Self-Cond-GAN + Random Labels~\cite{liu2020diverse}  & 
    % $240.0\mhide{\pm 12.02}$ & $2.90 \mhide{\pm 0.192}$ & ${29.04} \mhide{\pm 0.76}$ & ${6.97} \mhide{\pm 0.062}$ & ${0.0100} \mhide{\pm 0.0010}$ \\
    % Self-Cond-GAN+ Online Clustering~\cite{liu2020diverse}  & 
    % ${995.8} \mhide{\pm 0.86}$ & $0.17 \mhide{\pm 0.027}$ & ${31.56} \mhide{\pm 0.48}$ & ${6.82} \mhide{\pm 0.112}$ & ${0.0178} \mhide{\pm 0.0029}$ \\
    Self-Cond-GAN~\cite{liu2020diverse} & 
    $\textbf{1000.0} \mhide{\pm 0.00}$ & $0.08 \mhide{\pm 0.009}$ &  $18.03 \pm 0.55$ & $7.72 \pm 0.034$ & $0.0015 \mhide{\pm 0.0004}$ \\
    Random Partition ID &  $ 570 \mhide{\pm  14.05}$ & $  1.23 \mhide{\pm 0.352}$ & $ 22.57 \mhide{\pm 0.59}$ &  ${7.47} \mhide{\pm 0.051}$ & ${0.0074} \mhide{\pm 0.0008}$ \\
    Partition GAN (Ours) &   $\textbf{1000.0} \mhide{\pm 0.00}$ & $\textbf{0.02} \mhide{\pm 0.005}$ & $10.69 \mhide{\pm 0.33}$ &  ${8.52} \mhide{\pm 0.075}$ & $0.0005 \mhide{\pm 0.0002}$ \\ 
    PGMGAN (Ours) &  $\textbf{1000.0} \mhide{\pm 0.00}$ & $\textbf{0.02} \mhide{\pm 0.003}$& $\textbf{8.93} \mhide{\pm 0.38}$ &  $\textbf{8.81} \mhide{\pm 0.101}$ & $\textbf{0.0004} \mhide{\pm 0.0001}$ \\
    \midrule
    Logo-GAN-RC~\cite{sage2018logo} & 
    $\textbf{1000.0} \mhide{\pm 0.00}$ & $0.08 \mhide{\pm 0.006}$ &  $28.83 \mhide{\pm 0.43}$ &  ${7.12} \mhide{\pm 0.047}$ & ${0.0091} \mhide{\pm 0.0001}$ \\
    Class-conditional GAN~\cite{mirza2014conditional} & 
    $\textbf{1000.0} \mhide{\pm 0.00}$ & $0.08 \mhide{\pm 0.003}$ &   ${23.56} \mhide{\pm 2.24}$ & ${7.44} \mhide{\pm 0.080}$ & ${0.0019} \mhide{\pm 0.0001}$ \\
    \bottomrule
  \end{tabular}%
 }
\caption{Performance comparison of the unsupervised (above midline)/supervised (below midline) image generation methods on the Stacked MNIST and CIFAR-10 datasets. The number of recovered modes, reverse KL, FID, and IS are used as the evaluation metrics. We report the means and standard deviations over five random initializations. For CIFAR-10, all methods recover all 10 modes. Results of the compared models are quoted from Liu et al.~\cite{liu2020diverse}
}
\label{tab:stackedmnist_cifar_results}
\end{table*}

\section{Experiments}
\label{sec:experiments}

This section provides an empirical analysis of our method on various datasets\footnote{~The code to reproduce experiments is available at \protect\url{https://github.com/alisadeghian/PGMGAN}}. We adopt the architecture of SN-GAN \cite{miyato2018spectral} for our generators and discriminators. We use a Lipschitz constant of $0.9$ for our space partitioner that consists of 20 residual blocks, resulting in 60 convolutional layers. We use Adam optimizer~\cite{kingma2014adam} to train the proposed generators, discriminators, and space partitioner, and use SGD for training the pretext model.
%All experiments except the ImageNet are performed using two RTX 2080 Ti GPUs. 
Please refer to the Appendix for complete details of hyper-parameters and architectures used for each component of our model. \\

\textbf{Datasets and evaluation metrics} We conduct extensive experiments on CIFAR-10~\cite{krizhevsky2009learning} and STL-10~\cite{coates2011stl10} (48$\times$48), two real image datasets widely used as benchmarks for image generation. To see how our method fares against large dataset, we also applied our method on
ILSVRC2012 dataset (ImageNet) \cite{russakovsky2015imagenet} which %we compressed  to 128$\times$128 pixel
were downsampled to 128$\times$128$\times$3. To evaluate and compare our results, we use Inception Score (IS)~\cite{salimans2016improved} and Frechet Inception Distance (FID)~\cite{heusel2017gans}. It has been shown that IS may have many shortcomings, especially on non-ImageNet datasets. FID can detect mode collapse to some extent for larger datasets~\cite{borji2019pros, lucic2018gans, sajjadi2018assessing}. However, since FID is not still a perfect metric, we also evaluate our models using \emph{reverse-KL} which reflects both mode dropping and spurious~modes \cite{lucic2018gans}. All FIDs and Inception Scores (IS) are reported using 50k samples. No truncation trick is used to sample from the generator.

We also conduct experiments on three synthetic datasets: Stacked-MNIST~\cite{lin2018pacgan,liu2020diverse, srivastava2017veegan} that has up to 1000 modes, produced by stacking three randomly sampled MNIST~\cite{lecun2010mnist} digits into the three channels of an RGB image, and the 2D-grid dataset described in Section~\ref{sec:2d_experiments} as well as 2D-ring dataset. The empirical results of the two later datasets are presented in the Appendix.

\begin{figure}[t!]
\begin{subfigure}{.25\textwidth}
  \centering
%   \caption*{~~~~Real~~~~~~~~~~~~~Generated} \vspace{-8pt}
  \includegraphics[width=.8\linewidth]{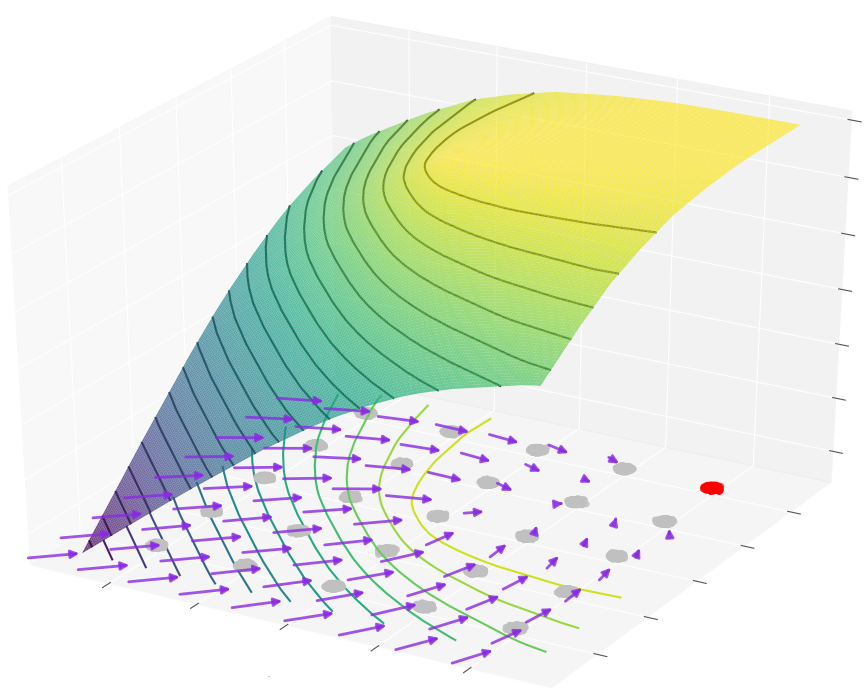}
%   \caption{PGMGAN}
%   \label{fig:2d_sfig3}
\end{subfigure}%
\begin{subfigure}{.25\textwidth}
  \centering
%   \caption*{~~~~Real~~~~~~~~~~~~~Generated} \vspace{-8pt}
\includegraphics[width=.8\linewidth]{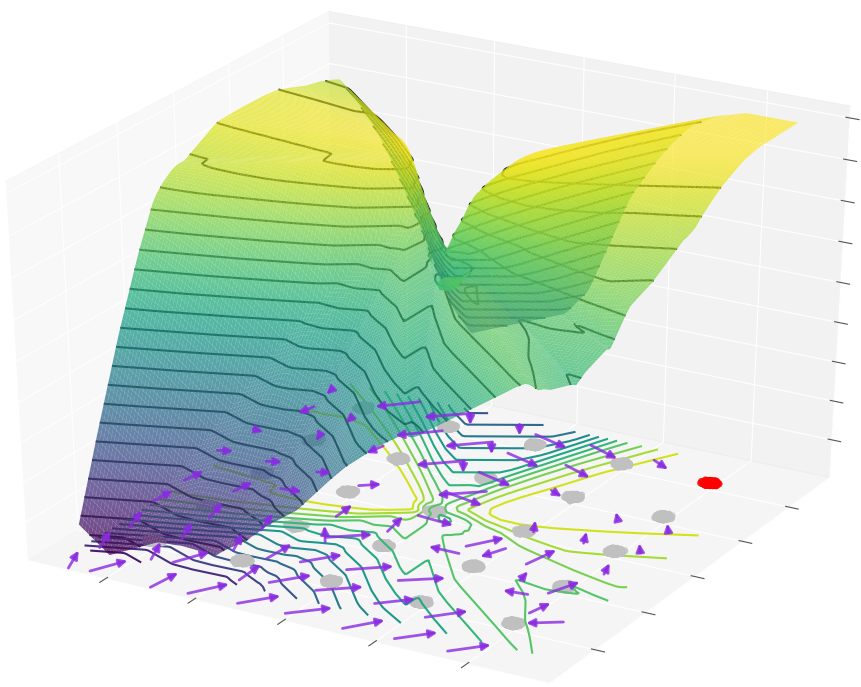}
%   \includegraphics[width=.98\linewidth]{images/2d_experiments/invertible/18.png}
%   \caption{invertible gan}
%   \label{fig:2d_sfig4}
\end{subfigure}
\caption{Left/right: the graph of $-R_i(\rvx)$ with/without assumption on the architecture, where the data points of the $i$\textsuperscript{th} partition are shown in red.}
\label{fig:2d_guide}
\end{figure}
\begin{figure}[t!]
\begin{subfigure}{.25\textwidth}
  \centering
%   \caption*{~~~~Real~~~~~~~~~~~~~Generated} \vspace{-8pt}
  \includegraphics[width=.8\linewidth]{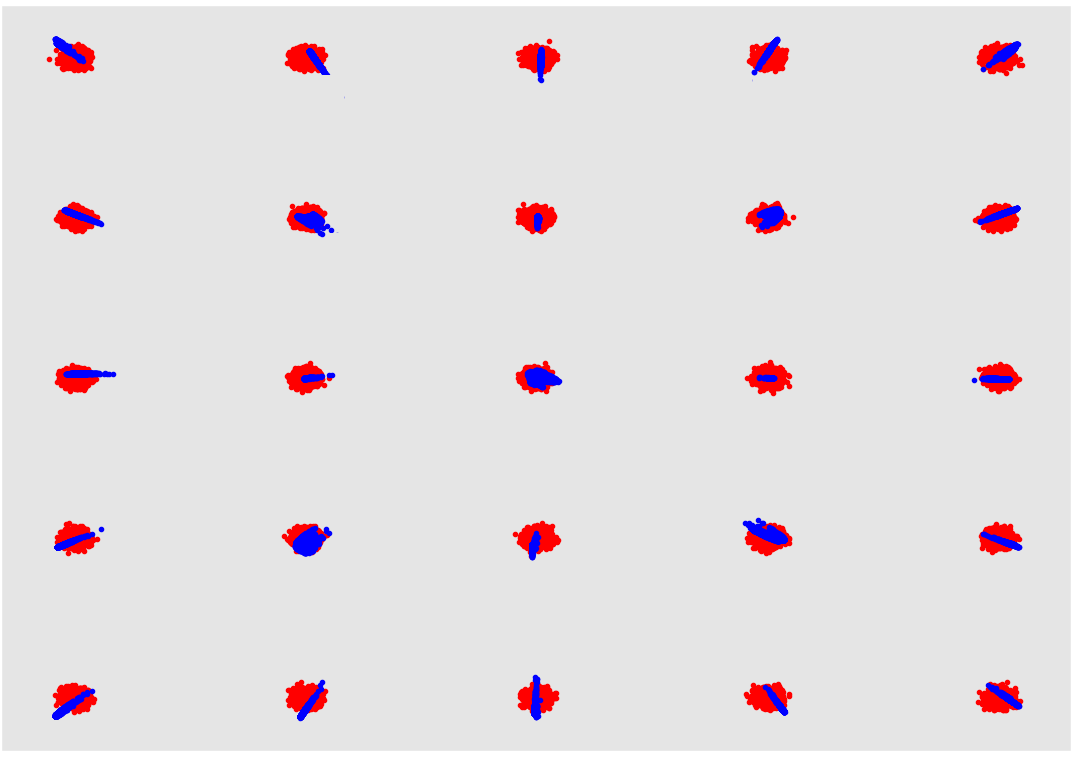}
%   \caption{Vanilla Partition GAN}
\end{subfigure}%
\begin{subfigure}{.25\textwidth}
  \centering
%   \caption*{~~~~Real~~~~~~~~~~~~~Generated} \vspace{-8pt}
  \includegraphics[width=.8\linewidth]{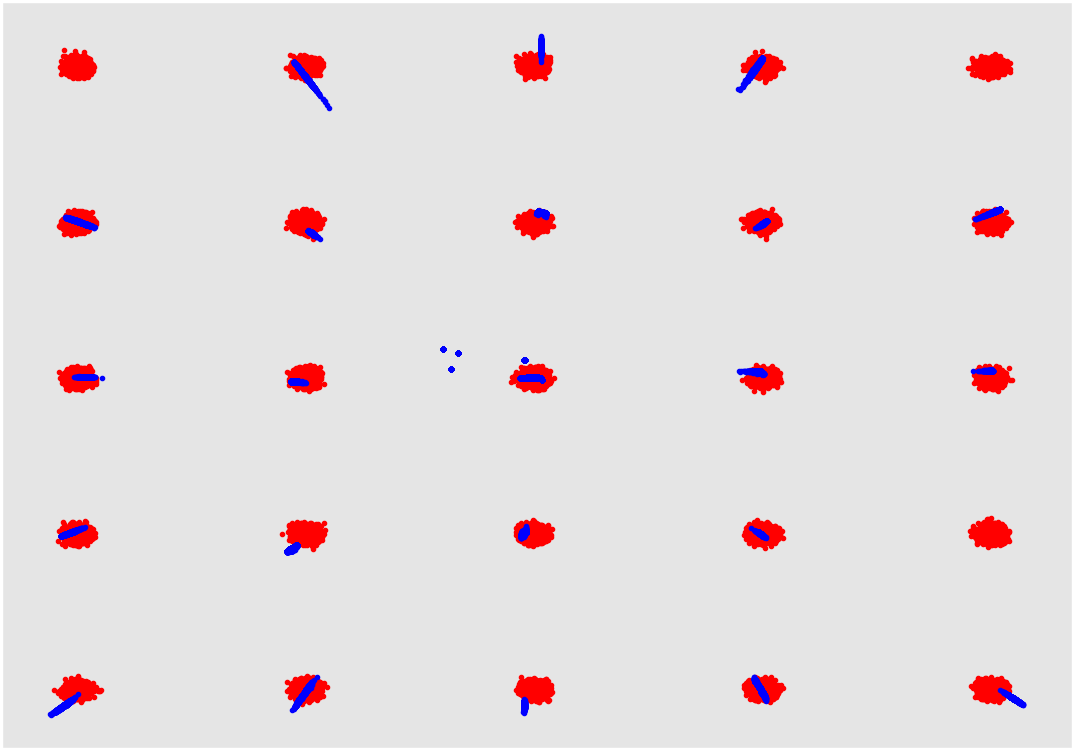}
%   \includegraphics[width=.48\linewidth]{images/2d_experiments/fullyconnected/7.png}
%   \caption{FC guide}
\end{subfigure}
\caption{Left/right: visual comparison of generated samples on the 2D-grid dataset using PGMGAN, with/without architecture restriction for the space partitioner. The red/blue points illustrate the real/generated data samples. In the right plot, some modes are missed and their corresponding generators focus on the wrong area.}
\label{fig:2d_gan}
\end{figure}

\subsection{Toy dataset}
\label{sec:2d_experiments}

This section aims to illustrate the importance of having a proper guide with no local optima. We also provide intuition about how our method helps GAN training. To that end, we use the canonical 2D-grid dataset, a mixture of 25 bivariate Gaussian with identical variance, and means covering the vertices of a square lattice. 

For this toy example the data points are low dimensional, thus we skip the feature learning step and directly train the space partitioner $S$. We train our space partitioner using two different architectures: one sets its neural network architecture to a multi-layer fully connected network with ReLU activations, while the other follows the properties of architecture construction in Section~\ref{sec:partition_guide_gan}. Once the two networks are trained, both successfully learn to put each Gaussian component in a different cluster, i.e., both get perfect clustering accuracy. Nonetheless, the \textit{guide} functions obtained from each architecture behave significantly different.

Figure~\ref{fig:2d_guide} provides the graph of $-R_i(\vx)$ for the two space partitioners, where $i$ is the partition ID for the red Gaussian data samples in the corner. The right plot shows that $-R_i(\rvx)$ can have undesired local optima when the conditions of Section~\ref{sec:partition_guide_gan} are not enforced. Therefore, a universal reliable gradient is not provided to move the data samples toward the partition of interest. On the other hand, when the guide's architecture follows these conditions (left plot), taking the direction of $\nabla-R_i(\vx)$ guarantees reaching to partition $i$. 

Figure~\ref{fig:2d_gan} shows the effect of both these guides in the training of our mixture of generators using Equation~\ref{eq:regguide}. As shown, when $R_i$ has local optima, the generator of that region may get stuck in those local optima and miss the desired mode. As shown in Liu et al.~\cite{liu2020diverse} and  the Appendix, GANs trained with no guide also fail to generate all the modes in this dataset. Furthermore, in contrast to standard GANs, we don't generate samples from the space between different modes due to our partitioning and mixture of generators, mitigating the mode connecting problem. We provide empirical results for this dataset in the Appendix. 

\subsection{Stacked-MNIST, CIFAR-10 and STL-10}
\label{sec:exp_guide}

In this section, we conduct extensive experiments to evaluate the proposed Partition-Guided Mixture of Generators (PGMGAN) model. We also quantify the performance gains for the different parts of our method through an ablation study. We randomly generated the partition labels in one baseline to isolate the effect of proper partitioning from the architecture choice of $G_i$/$D_i$'s. We also ablate the benefits of the guide function by making a baseline where $\lambda=0$. For all experiments, we use $k=200$ unless~specified~otherwise. 

Tables~\ref{tab:stackedmnist_cifar_results} and~\ref{tab:stl_results} presents our results on Stacked MNIST, CIFAR-10 and STL-10 respectively. From these tables, it is evident how training multiple generators using the space partitioner allows us to significantly outperform the other benchmark algorithms in terms of all metrics. Comparing \textit{Random Partition ID} to \textit{Partition+GAN} clearly shows the importance of having an effective partitioning in terms of performance and mode covering. Furthermore, the substantial gap between \textit{PGMGAN} and \textit{Partition+GAN} empirically demonstrates the value of utilizing the guide term. 

\begin{table}[h]
\centering%
\caption{Unsupervised image generation results on STL-10. The results of all compared method are taken from Tian et al. \cite{tian2020off}}
\label{tab:stl_results}
\vspace{-3mm}
\resizebox{%
  \ifdim\width>\columnwidth
    \columnwidth
  \else
    0.70\columnwidth
  \fi
}{!}
{%
\begin{tabular}{lrr}
\toprule
\multicolumn{1}{c}{} &
\multicolumn{2}{c}{STL-10} \\
\cmidrule(r){2-3}
        & \multicolumn{1}{c}{FID $\downarrow$}  &  \multicolumn{1}{c}{IS $\uparrow$} \\
\midrule
D2GAN~\cite{nguyen2017dual}               & -   & 7.98 \\
DFM~\cite{warde2016improving}              & - & 8.51  \\
ProbGAN~\cite{he2019probgan}              & 46.74   & 8.87  \\
SN-GAN~\cite{miyato2018spectral}         & 40.15   & 9.10  \\
Dist-GAN~\cite{tran2018dist}              & 36.19  &  -   \\
MGAN~\cite{hoang2018mgan}                 & -    & 9.22 \\
Improved MMD~\cite{wang2018improving}     & 37.63  & 9.34  \\
AGAN~\cite{wang2019agan}                 & 52.75   & 9.23  \\
AutoGAN~\cite{gong2019autogan}            & 31.01   & 9.16 \\
E$^2$GAN~\cite{tian2020off}                & 25.35 & 9.51  \\
Partition GAN (Ours)                      & 26.28  & 10.35  \\
PGMGAN (Ours)                     & \textbf{19.52} & \textbf{11.16}\\
\bottomrule
\vspace{-18pt}
\end{tabular}
} %
\vspace{10pt}
\end{table}

We first perform overall comparisons against other recent GANs. As shown in Table~\ref{tab:stl_results}, PGMGAN achieves state-of-the-art FID and IS on the STL-10 dataset. Furthermore, PGMGAN outperforms several other baseline models, even over supervised class-conditional GANs, on both CIFAR-10 and Stacked-MNIST, as shown in Table~\ref{tab:stackedmnist_cifar_results}. The significant improvements of FID and IS reflect the large gains in diversity and image quality on these datasets.

Following Liu et al.~\cite{liu2020diverse}, we calculate the reverse KL metric using pre-trained classifiers to classify and count the occurrences of each mode for both Stacked-MNIST and CIFAR-10. These experiments and comparisons demonstrate that our proposed guide model effectively improves the performance of GANs in terms of mode~collapse.

\subsection{Image generation on unsupervised ImageNet}
To show that our method remains effective on a larger more complex dataset, we also evaluated our model on unsupervised ILRSVRC2012 (ImageNet) dataset which contains roughly 1.2 million images with 1000 distinct categories; we down-sample the images to 128$\times$128 resolution for the experiment. We use $k=1000$ and adopt the architecture of BigGAN \cite{brock2018large} for our generators and discriminators.
Please see the Appendix for the details of experimental settings. 

Our results are presented in Table \ref{tab:imagenet}. To the best of our knowledge, we achieve a new state of the art (SOTA) in unsupervised generation on ImagNet apart from augmentation based methods.

\begin{table}[h]
  \centering%
   \caption{FID and Inception Score (IS) metrics for unsupervised image generation on ImageNet at resolution 128$\times$128. The results of all compared methods are taken from Liu et al. \cite{liu2020diverse}} 
   \label{tab:imagenet}
    \vspace{-3mm}
    \resizebox{%
0.8\columnwidth
}{!}
    {%
    
    \begin{tabular}{lrr}
    \toprule
    \multicolumn{1}{c}{} &
    \multicolumn{2}{c}{ImageNet} \\
    \cmidrule(r){2-3}
            & \multicolumn{1}{c}{FID $\downarrow$}  & \multicolumn{1}{c}{IS $\uparrow$}  \\
    \midrule
    GAN~\cite{goodfellow2014generative}   & 54.17  & 14.01  \\
    PacGAN2~\cite{lin2018pacgan}   & 57.51  & 13.50  \\
    PacGAN3~\cite{lin2018pacgan}  & 66.97  & 12.34  \\
    \mhide{MGAN~\cite{hoang2018mgan}}  & \mhide{58.88}  & \mhide{13.22}  \\
    \mhide{RotNet Feature Clustering}   & \mhide{53.75}  & \mhide{13.76}  \\
    Logo-GAN-AE~\cite{sage2018logo}   & 50.90  & 14.44  \\
    Self-Cond-GAN~\cite{liu2020diverse}   & 40.30 & 15.82  \\
    PGMGAN (Ours) & \textbf{21.73} & \textbf{23.31}  \\
    \bottomrule
  \end{tabular}
  } %
  \vspace{-3pt}
\end{table}
%We observe that our method generates both higher quality and more diverse samples than competing unsupervised methods.

\subsection{Parameter sensitivity}
\label{sec:ablation_exp}

\begin{figure}[h]
\centering
\caption{Effect of changing the guide's weight $\lambda$ in equation~\ref{eq:regguide} on PGMGAN performance. $\lambda=0$ corresponds to the partition+GAN.}
\includegraphics[width=0.28\textwidth]{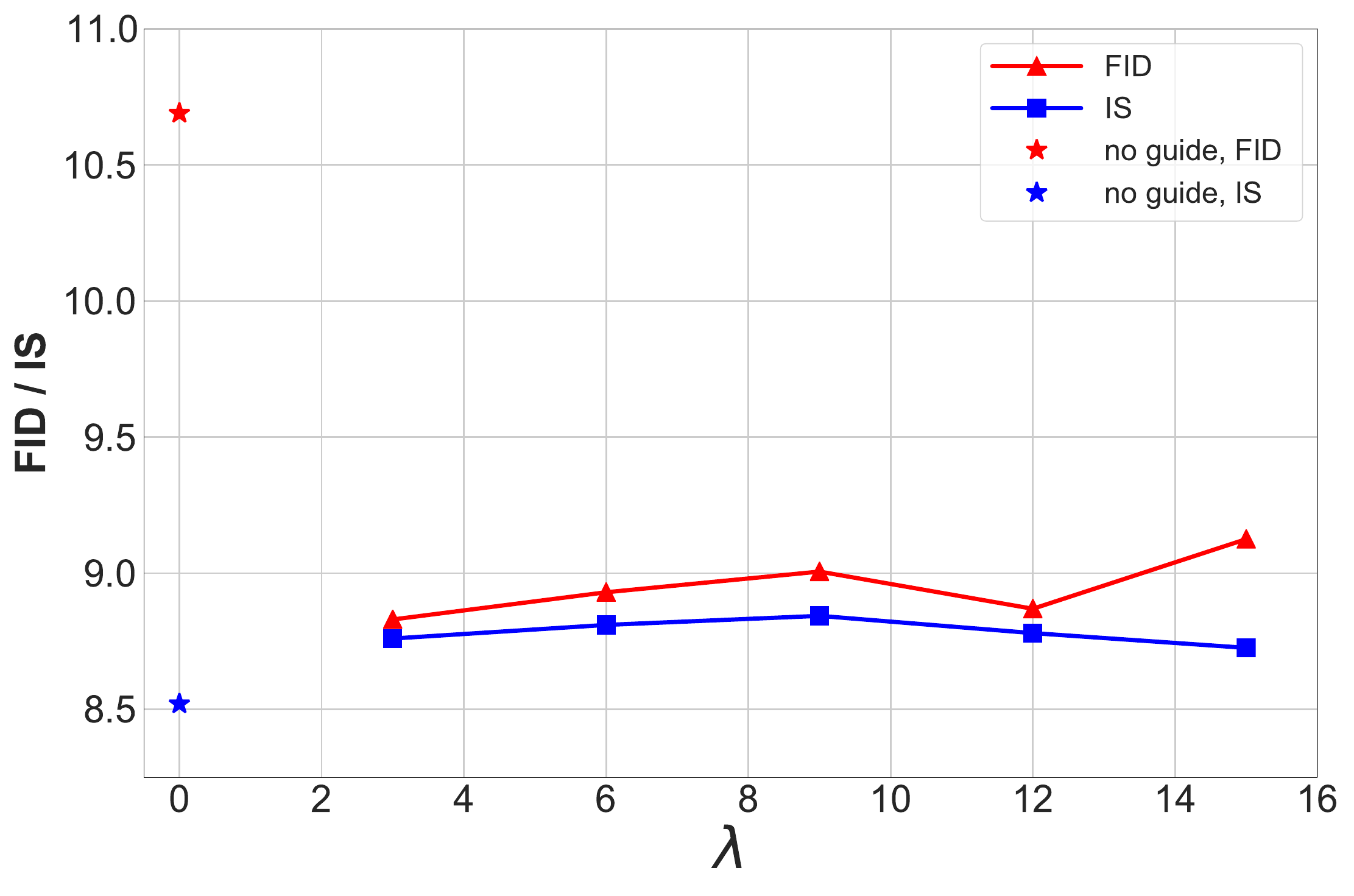}
\label{fig:sensitivity_lambda}
\end{figure}

Additionally, we study the sensitivity of our overall \mbox{PGMGAN} method to the choice of guide's weight $\lambda$ (Eq.~\ref{eq:regguide}), and number of clusters $k$. Figure~\ref{fig:sensitivity_lambda} shows the results with varying $\lambda$ over CIFAR-10, demonstrating that our method is relatively robust to the choice of this hyper-parameter. Next, we change $k$ for a fixed $\lambda = 6.0$ and report the results in Table~\ref{tab:effect_k_cifar}. we observe that our method performs~well~for~a~wide~range of~$k$.

\begin{table}[h]
\centering%
\caption{Effect of number of partitions ($k$) on PGMGAN performance. Results are averaged over five random trials, with standard error reported.}
\label{tab:effect_k_cifar}
\vspace{-3mm}
\resizebox{%
    0.8\columnwidth
}{!}
{%
\begin{tabular}{lrr}
\toprule
\multicolumn{1}{c}{} &
\multicolumn{2}{c}{CIFAR-10} \\
\cmidrule(r){2-3}
        & \multicolumn{1}{c}{FID $\downarrow$}  &  \multicolumn{1}{c}{IS $\uparrow$} \\
\midrule
GAN~\cite{goodfellow2014generative} &  $28.08 \mhide{\pm 0.47}$ &  ${6.98} \mhide{\pm 0.06}$ \\
PGMGAN ($k=50$)  &  ${9.27} \mhide{\pm  0.45}$ & ${8.69} \pm  0.07$ \\
PGMGAN ($k=100$) &  ${8.97} \mhide{\pm  0.33}$ & ${8.75} \pm 0.05$ \\
PGMGAN ($k=200$) &  ${8.93} \mhide{\pm  0.38}$ & ${8.81} \pm 0.10$ \\
% $k=1000$  &  ${20.76} \pm 0.26$ & ${7.48} \pm 0.028$ \\
\midrule
Class Conditional GAN~\cite{mirza2014conditional} & ${23.56} \mhide{\pm 2.24}$ & ${7.44} \mhide{\pm 0.08}$ \\
\bottomrule
\vspace{-5pt}
\end{tabular}
} %
% \vspace{10pt}
\end{table}

\vspace{-10pt}
\section{Conclusion}

We introduce a differentiable space partitioner to alleviate the GAN training problems, including mode connecting and mode collapse. The intuition behind how this works is twofold. The first reason is that an efficient partitioning makes the distribution on each region simpler, making its approximation easier. Thus, we can have a better approximation as a whole, which can alleviate both mode collapse and connecting problems. The second intuition is that the space partitioner can provide extra gradient, assisting the discriminator in training the mixture of generators. This is especially helpful when the discriminator's gradient is unreliable. However, it is crucial to have theoretical guarantees that this extra gradient does not deteriorate the GAN training convergence in any way. We identify a sufficient theoretical condition for the space partitioner (in the functional space), and we realize that condition empirically by an architecture design for the space partitioner.  Our experiments on natural images show the proposed method improves existing ones in terms of both FID and IS.
For future work, we would like to investigate using the space partitioner for the supervised regime, where each data label has its own partitioning. Designing a more flexible architecture for the space partitioner such that its guide function does not have local optima is another direction we hope to explore.

\section*{Acknowledgments.} 
This work was supported in part by NSF
%The authors acknowledge the support of Grant 
IIS-1812699. % from the U.S. National Science Foundation.

\clearpage

{\small
\bibliographystyle{ieee_fullname}
\bibliography{ref}
}
\clearpage

\begin{appendices}
\pagestyle{empty}
\section{Proofs}
\label{sec:app_proofs}

\ModeConnectingThm*
\begin{proof} We begin by re-stating the definition of Minkowski sum and proceed by proving the theorem for the case where the number of manifold is 2. To extend the theorem form~\mbox{$k=2$} to the general case, one only needs to consider manifold $\mathcal{M}_i$ as $\mathcal{M}_1$, and $\bigcup_{j \in [1:k]\setminus {i}}\mathcal{M}_j$ as $\mathcal{M}_2$.

\begin{restatable}[Minkowski sum]{defi}{minkowski_sum}
\label{def:minkowski_sum}
The Minkowski sum of two sets $U, V \in \sR^d$ defined as
$$U+V := \{u+v | u \in U, v \in V \}$$
and when $V$ is a $d$ dimensional ball with radius $r$ and centered at zero, we use the notation $U_r$ to refer to their Minkowski sum. 
\end{restatable}

% \begin{definition}[Minkowski sum]
% The Minkowski sum of two sets $U, V \in \sR^d$ defined as
% $$U+V := \{u+v | u \in U, v \in V \}$$
% and when $V$ is a $d$ dimensional ball with radius $r$ and centered at zero, we use the notation $U_r$ to refer to their Minkowski sum. 
% \end{definition}

\noindent If we let $U^{(1)} := G_{\theta}^{-1}(\mathcal{M}_1), U^{(2)} := G_{\theta}^{-1}(\mathcal{M}_2)$, then: 
$$\forall r_1,r_2 \in \sR_+,~\text{if}~~~ r_1 + r_2 < d_1/c \implies U^{(1)}_{r_1} \cap U^{(2)}_{r_2} = \emptyset $$
that is because if there exists an $\rvx \in U^{(1)}_{r_1} \cap U^{(2)}_{r_2} $, there would be $\rvu_1 \in U^{(1)}, \rvu_2 \in U^{(2)}$ such that:
\begin{equation}
    \resizebox{0.91\hsize}{!}{$||\rvx - \rvu_1||\leq r_1, ~~~ ||\rvx - \rvu_2||\leq r_2 \implies ||\rvu_1 - \rvu_2||<r_1+r_2< d_1/c$}
\end{equation}

\noindent However, due to lipsitz condition of  $G_{\theta}$:  $$||G_{\theta}(\rvu_1) - G_{\theta}(\rvu_2)||< c||\rvu_1 - \rvu_2||< c \cdot d_1/c =d_1 $$
which contradicts with our assumption that the distance between $\mathcal{M}_1,\mathcal{M}_2$ is $d_1$. Therefore there is no point in the intersection of $U^{(1)}_{r_1}$ and $U^{(2)}_{r_2}$. The disjointness of this two sets provides us:
$$\gamma_n(U^{(1)}_{r_1}) + \gamma_n(U^{(2)}_{r_2}) \leq \gamma_n(\sR^n)=1 $$
where $\gamma_n(.)$ of any set is the probability of a random draw of $\mathcal{N}(0, \mI_n)$ being from that set.
We proceed by using a remark from theorem 1.3 of \cite{ledoux1996isoperimetry} which restated below:

\begin{lemma}
If $U$ is a Borel set in $\sR^n$, then:
$$p \leq \gamma_n(U) \implies \Phi( \Phi^{-1}(p) + r) \leq \gamma_n(U_r).$$
\end{lemma}

\noindent Based on above lemma if we let:
$$p_1 := \gamma_n(U^{(1)}), \;\; p_2:= \gamma_n(U^{(2)})$$
then for $\forall r_1,r_2 \in \sR_+$ such that $r_1 + r_2 < d_1/c$, we have:
\begin{equation}\label{eq:lessthanone}
    \Phi( \Phi^{-1}(p_1) + r_1) +\Phi( \Phi^{-1}(p_2) + r_2)  \leq 1
\end{equation}

\noindent We can now calculate the total variational distance of the marginal distributions of $\pdata,\, \pmodel$ on the set $\{ G_{\theta}(U^{(1)}), G_{\theta}(U^{(2)}), G_{\theta}\left ( \sR^n \setminus (U^{(1)} \cup U^{(2)}) \right ) \}$ as:
\begin{equation}
\label{eq:dtv}
\resizebox{0.91\hsize}{!}{$
  d_{TV}(\pdata^{(\textit{marginal})}, \pmodel^{(\textit{marginal})}) = |\pi_1 -p_1| + |\pi_2 -p_2| + |1- (p_1+p_2)|$} 
\end{equation}

\noindent and since total variational distance takes a smaller value on marginal distributions than the full distribution, we only need to show that for any $i$ the expression in the equation \ref{eq:dtv} is larger than $g(\pi_i, d_i, c)$ to prove the theorem \ref{thm:mode_connecting} for $k=2$. Here $g(\pi_i, d_i, c) = \pi_i^* - \Phi( \Phi^{-1}(\pi_i^*) - d_i/c)$. \\

Assume $p_1 \leq \pi_1$, and define~$\Delta_1 : =\pi_1 - p_1\geq0$, based on equation \ref{eq:lessthanone} if $r_1=d_1/c, \,\; r_2=0$, we have:
$$ \Phi( \Phi^{-1}(\pi_1- \Delta_1) + d_1/c) + p_2  \leq 1$$
which based on equation \ref{eq:lessthanone} implies:
$$r(\Delta_1):=\Phi( \Phi^{-1}(\pi_1- \Delta_1)+d_1/c) - (\pi_1- \Delta_1) \leq D_{TV}$$
which $D_{TV}$ refers to total variational distance between the marginal distributions of the data and model.
We also know from the equation \ref{eq:lessthanone}, that
$ \Delta_1 \leq D_{TV}$, therefore:
$$\max(r(\Delta_1), \Delta_1) \leq D_{TV}$$
for a $\Delta_1 \in [0, \pi_1]$. Therefore 
$$\min_{\delta_1 \in [0, \pi_1]}\{\max(r(\delta_1), \delta_1)\} \leq D_{TV} $$
To find the $\delta_1$ which minimize the above equation, we need to check endpoints of the interval $[0, \pi_1]$, points where the curve of two functions $r(\delta_1), \delta_1$ intersects with each other, and points that are the local minimaum of each of them. It can be shown the function $r(\delta_1)$ does not have any local minima when $0<\pi_1<1$ because:
\begin{equation}
  r(\delta_1)= P(\rz \in [\Phi^{-1}(\pi_1- \delta_1), \Phi^{-1}(\pi_1- \delta_1)+d_1/c] )  
\end{equation}
where $\rz$ is univarite standard normal random variable. Therefore $r(\delta_1)$ is the probablity of a univarite normal being in a fixed length interval $d_1/c$, and $\delta_1$ only changes the starting point of the interval. By using this fact, it can be easily shown this function does not have any local optima in the open interval $(0, \pi_1)$. Also the identity function $\delta_1$ also has no local optima inside the interval. The endpoints values~are:
$$\max(r(0), 0)= \Phi( \Phi^{-1}(\pi_1)+d_1/c) - \pi_1$$
$$\max(r(\pi_1), \pi_1)= \max(\Phi( \Phi^{-1}(0)+d_1/c), \pi_1)= \pi_1 $$

\noindent The function curves of $r$ and identity also intersects only when:
$$\Phi( \Phi^{-1}(\pi_1- \delta_1^*)+d_1/c) = \pi_1 $$
which only happens when:
$$ \pi_1 - \Phi( \Phi^{-1}(\pi_1) - d_1/c) = \delta_1^* $$
where for this point,
$\max(r(\delta_1^*), \delta_1^*) =\delta_1^*$. 
%g(\pi_i, d_i, c) &:=
\noindent Therefore based on the above calculations:
\begin{dmath*}
\min \left \{\underbrace{\Phi( \Phi^{-1} (\pi_1)+  d_1/c) - \pi_1}_{\text{I}}\,, \, \pi_1, \\
\underbrace{\pi_1 - \Phi( \Phi^{-1}(\pi_1) - d_1/c)}_{\text{II}} \right \} \leq D_{TV}
\end{dmath*}

Note, $\pi_1$ is always smaller than term II, and term I (II) is equal to probability of a univariate standard normal random variable being inside the interval $[\Phi^{-1}(\pi_1), \Phi^{-1}(\pi_1) + d_1/c]$ ($[\Phi^{-1}(\pi_1)- d_1/c, \Phi^{-1}(\pi_1) ]$). This observation implies that term II is smaller than term I, if and only if $\pi_1 \leq \pi_2$. Based on this fact and symmetry of $\Phi$ with respect to zero, it can be easily shown that:
\begin{equation}\label{eq:lastproof}
  g(\pi_1, d_1, c)\leq D_{TV} 
\end{equation}
which proves the theorem. However, we made an assumption that $p_1\leq \pi_1$, this does not harm the argument because otherwise we would have $p_2\leq \pi_2$, and we can restate all the above arguments for $\pi_2$ instead of $\pi_1$. And, since $\pi_2=1-\pi_1$ and $d_1=d_2$, therefore we can have $g(\pi_1, d_1, c)= g(\pi_2, d_2, c)$, which proves equation \ref{eq:lastproof}.
\end{proof}

% \clearpage

\MixtureThm*
\begin{proof}
Based on the definition of the JSD, we have:
\begin{align*}
    {\rm JSD}(P \parallel Q) =\frac{1}{2} {\rm KL}(P \parallel  \frac{P+Q}{2}) +\frac{1}{2} {\rm KL}( Q \parallel \frac{P+Q}{2} )
\end{align*}
We also have:
\begin{align*}
    {\rm KL}(P \parallel & \frac{P+Q}{2})  =\\
    \int_{\sR^d} & P(x) \log \frac{P(x)}{P(x)+ Q(x)} \, dx  + \log 2 =\\
   \sum_i  \int_{\sA_i} & P(x) \log \frac{P(x)}{P(x)+ Q(x)} \, dx   + \log 2=\\
   \sum_i  \int_{\sA_i} & \pi_i p_i(x) \log \frac{\pi_i p_i(x)}{\pi_i p_i(x)+ \pi_i q_i(x)} \, dx   + \log 2=\\
    \sum_i \pi_i & (\int_{\sA_i}  p_i(x) \log \frac{ p_i(x)}{ p_i(x)+  q_i(x)} \, dx   + \log 2)=\\
    &\sum_i \pi_i  {\rm KL}(p_i  \parallel  \frac{p_i+q_i}{2}).
\end{align*}
Therefore:
\begin{align*}
    {\rm KL}(P \parallel  \frac{P+Q}{2})  = \sum_i \pi_i  {\rm KL}(p_i  \parallel  \frac{p_i+q_i}{2}),
\end{align*}
and similarly:
\begin{align*}
    {\rm KL}(Q \parallel  \frac{P+Q}{2}  )  =
    \sum_i \pi_i  {\rm KL}( q_i \parallel  \frac{p_i+q_i}{2} )
\end{align*}
% and by adding these two terms, we have:
% \begin{align*}
%     {\rm JSD}(P \parallel Q) =  \frac{1}{2}  {\rm KL}(P \parallel & \frac{P+Q}{2}) +\frac{1}{2} {\rm KL}(Q \parallel  \frac{P+Q}{2}  )=\\
%     \frac{1}{2} \sum_i \pi_i  {\rm KL}(p_i  \parallel  \frac{p_i+q_i}{2}) + &
%     \frac{1}{2} \sum_i \pi_i  {\rm KL}( q_i \parallel  \frac{p_i+q_i}{2} )=\\
%     \sum_i \frac{\pi_i}{2} ( {\rm KL}(p_i  \parallel  \frac{p_i+q_i}{2}) + & 
%   {\rm KL}( q_i \parallel  \frac{p_i+q_i}{2} ) )=\\
%   \sum_i \pi_i {\rm JSD}(p_i \parallel q_i)
% \end{align*}

% The above completes the proof of theorem. 
\noindent Adding these two terms completes the proof.
\end{proof}

% \newpage

% \begin{align*}
%     {\rm JSD}(P \parallel Q)  = &\\ \frac{1}{2}  {\rm KL}(P  \parallel  \frac{P+Q}{2}) & +\frac{1}{2} {\rm KL}(Q \parallel  \frac{P+Q}{2}  )=\\
%     \frac{1}{2} \sum_i \pi_i  {\rm KL}(p_i  \parallel  \frac{p_i+q_i}{2}) + &
%     \frac{1}{2} \sum_i \pi_i  {\rm KL}( q_i \parallel  \frac{p_i+q_i}{2} )=\\
%     \sum_i \frac{\pi_i}{2} ( {\rm KL}(p_i  \parallel  \frac{p_i+q_i}{2}) + & 
%   {\rm KL}( q_i \parallel  \frac{p_i+q_i}{2} ) )=\\
%   \sum_i \pi_i {\rm JSD}(p_i \parallel q_i) &
% \end{align*}

% \begin{dmath*}
%     {\rm JSD}(P \parallel Q) =\frac{1}{2}  {\rm KL}(P \parallel  \frac{P+Q}{2}) +\frac{1}{2} {\rm KL}(Q \parallel  \frac{P+Q}{2})=\\
%     \frac{1}{2} \sum_i \pi_i  {\rm KL}(p_i  \parallel  \frac{p_i+q_i}{2}) + 
%     \frac{1}{2} \sum_i \pi_i  {\rm KL}( q_i \parallel  \frac{p_i+q_i}{2} )=\\
%     \sum_i \frac{\pi_i}{2} ( {\rm KL}(p_i  \parallel  \frac{p_i+q_i}{2}) +  
%   {\rm KL}( q_i \parallel  \frac{p_i+q_i}{2} ) )=\\
%   \sum_i \pi_i {\rm JSD}(p_i \parallel q_i)
% \end{dmath*}

% \label{thm:mixture}
% Let $P = \sum_i^k \pi_i p_i$ , $Q = \sum_i^k \pi_i q_i$ , and $\sA_1, \sA_2,..., \sA_K$ be a partitioning of the space, such that the support of each distribution $p_i$ and $q_i$ is $\sA_i$. Then:\\
% \begin{equation}
% \label{eq:theorem1}
% {\rm JSD}(P \parallel Q) = \sum_i \pi_i {\rm JSD}(p_i \parallel q_i)
% \end{equation}

% \clearpage

\SuffThm*

\begin{proof}
We start by proving that the Jacobian matrix of function $\phi$ is invertible for any $\rvx \in \sR^d$. Since $\phi \in C^1$, based on Taylor's expansion theorem for multi-variable vector-valued function $\phi$, we can write:
$$\phi(\rvy)- \phi(\rvx) = \mJ_{\phi}(\rvx) \left(\rvy -\rvx \right) + o\left(||\rvy -\rvx||\right)$$
Were $o(\cdot)$ is the Little-o notation.
By taking norm from both sides and using triangle inequality, we have:
$$||\phi(\rvy)- \phi(\rvx) || \leq||\mJ_{\phi}(\rvx) \left(\rvy -\rvx \right)|| + ||o\left(||\rvy -\rvx||\right)||$$
Also because:
$$c_0 ||\rvy -\rvx||\leq ||\phi(\rvy)- \phi(\rvx) ||$$
\begin{equation}\label{eq:ineq}
    \implies c_0 ||\rvy -\rvx||\leq||\mJ_{\phi}(\rvx) \left(\rvy -\rvx \right)|| + ||o\left(||\rvy -\rvx||\right)||
\end{equation}

\noindent thus for any fixed $\rvx$:
$\exists \, \epsilon>0$ such that $\forall \, \rvy \in \sR^d$ where $||\rvy -\rvx||\leq \epsilon$ then:
$$ ||o\left(||\rvy -\rvx||\right)||\leq \frac{c_0}{2}||\rvy -\rvx||$$
which combined with the inequality \ref{eq:ineq}, results in:
$$\frac{c_0}{2} ||\rvy -\rvx||\leq||\mJ_{\phi}(\rvx) \left(\rvy -\rvx \right)||$$
For $\rvy \neq \rvx$, let $\rvu:= (\rvy -\rvx)/||\rvy -\rvx||$, then by dividing both sides of the above inequality to $||\rvy -\rvx||$ we have:
$$\forall \rvu \in \sR^d, ||\rvu||=1 \implies \frac{c_0}{2} \leq||\mJ_{\phi}(\rvx) \rvu||$$
which shows the Jacobian matrix of $\phi$ is invertiable for any $\rvx$ and all of its singular values are larger than $c_0/2$. If there is no $\rvx\in \sR^d\setminus A_i$ the proof is complete. Otherwise, consider any $\rvx \in \sR^d\setminus A_i$, for this $\rvx$ we have: 

$$0<R_i(\rvx) = \sum_c (f_c(\rvx) - f_i(\rvx))_{+}= \sum_c ((\vw_c- \vw_i)\phi(\rvx))_{+}$$
where $\vw_j$ is the $j$'th row of the matrix $\mW^{\textit{partitioner}}$.  Let: 
$$I(\rvx):= \{c | f_c(\rvx) > f_i(\rvx), ~~~ c\in [1:k]\}$$ 
which is an non-empty set, because $0<R_i(\rvx)$ and we have
\begin{align*}
0<R_i(\rvx)=\left[ \sum_{c\in I(\rvx)}(\vw_c- \vw_i)\right]\phi(\rvx) \\\implies \vv :=\sum_{c\in I(\rvx)}(\vw_c- \vw_i)\neq 0
\end{align*}

\noindent Taking the gradient of the new formulation of $R_i$, we have:
$$\nabla R_i(\rvx) =  \left[ \sum_{c\in I(\rvx)}(\vw_c- \vw_i)\right] \nabla(\phi(\rvx))= \vv \mJ_{\phi}(\rvx) $$

\noindent but since we showed earlier that all of the singular values of the Jacobian matrix is larger than $c_0/2$, the Jacobian matrix is $d\times d$, and $\vv$ is not equal to zero, it can be easily shown:
$$||\nabla R_i(\rvx)||> ||\vv||\frac{c_0}{2}:=b_0$$

Now, we also need to show $\sA_i$ is connected for any $i$ to complete the proof. To that end, we first show $\phi$ is a surjective function, which means its image is $\sR^d$. To show the $\phi$ is surjective, we prove its image is both an open and closed set, then since the only sets which are both open and closed (in $\sR^d$) are $\sR^d, \emptyset$, we can conclude the surejective property.  The image of $\phi$ is an open set due to Inverse Function Theorem~\cite{rudin1964principles} for $\phi$. We are allowed to use Inverse Function Theorem, since $\phi$ satisfies both $C^1$ condition and non zero determinant for all the points in the domain.
% (because we showed earlier all the singular values of the Jacobin are larger than $c_0/2$).
We will also show that the image of $\phi$ is a closed set by showing it contains all of its limit points. Let $ \vy$ be a limit point in the image of $\phi$, that is there exists $\{\vx_1, \vx_2, \cdots \}$ such that $\phi(\vx_r) \to \vy $. Since $\sR^d$ is complete and we have $c_0 ||\vx_r- \vx_s||\leq ||\phi(\vx_r)- \phi(\vx_s)||$, then $\{\vx_1, \vx_2, \cdots \}$ is a Cauchy sequence. Finally since $\phi$ is a continues function $\phi(\vx^*) =\vy$, completing the proof.
% The image of $\phi$ is also a closed set, because if $\phi(\vx_r) \to \vy $, it implies there exists $\vx^*$ in the domain of $\phi$, such that $\vx_r \to \vx^*$. That is because $\vx_1, \vx_2, \cdots$ is a    $c_0 ||\vx_r- \vx_s||\leq ||\phi(\vx_r)- \phi(\vx_s)||$ and completeness of the space $\sR^d$. And since $\phi$ is a continues function $\phi(\vx^*) =\vy$. Therefore, we completed the proof for $\phi$ being a surjective function.

The function $\phi$ is also an invertible function because if $\phi(\vx) = \phi(\vy)$ then 
$$c_0 ||\rvx - \vy|| \leq ||\phi(\vx) - \phi(\vy)|| = 0$$
which implies $\vx = \vy$. Therefore $\phi$ is in fact a continuous bijecitve function, which means it has a continuous inverse defined on all the space $\sR^d$. Furthermore, it can be easily shown $R_i$ for a datapoint is zero iff its transformation by $\phi$ lies in a polytope (where each of its facets is a hyperplance perpendicular to a $\vw_c- \vw_i$). Since convex polytope is a connected set, and by applying $\phi^{-1}$ (it is well defined everywhere because of bijective property of $\phi$) to it, we would have a connected set. That is because a continuous function does not change the connectivity and $\phi^{-1}$ is continuous. \end{proof}

\clearpage

\section{Additional qualitative results}
\label{sec:app_qualitative_results}

\begin{figure*}[b]
 \center
\begin{subfigure}{.29\textwidth}
  \centering
  \caption*{~~~~Real~~~~~~~~~~~~~~~~~~~~Generated} \vspace{-6pt}
  \includegraphics[width=.48\linewidth]{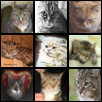}
  \includegraphics[width=.48\linewidth]{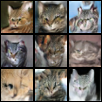}
  \caption{Partition 1}
  \label{fig:apx_cifar_sfig1}
\end{subfigure}
\begin{subfigure}{.29\textwidth}
  \centering
  \caption*{~~~~Real~~~~~~~~~~~~~~~~~~~~Generated} \vspace{-6pt}
  \includegraphics[width=.48\linewidth]{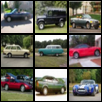}
  \includegraphics[width=.48\linewidth]{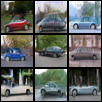}
  \caption{Partition 3}
  \label{fig:apx_cifar_sfig2}
\end{subfigure}
\begin{subfigure}{.29\textwidth}
  \centering
  \caption*{~~~~Real~~~~~~~~~~~~~~~~~~~~Generated} \vspace{-6pt}
  \includegraphics[width=.48\linewidth]{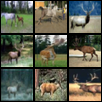}
  \includegraphics[width=.48\linewidth]{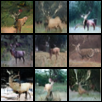}
  \caption{Partition 12}
  \label{fig:apx_cifar_sfig3}
\end{subfigure}
\begin{subfigure}{.29\textwidth}
  \centering
  \caption*{~~~~Real~~~~~~~~~~~~~~~~~~~~Generated} \vspace{-6pt}
  \includegraphics[width=.48\linewidth]{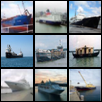}
  \includegraphics[width=.48\linewidth]{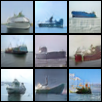}
  \caption{Partition 13}
  \label{fig:apx_cifar_sfig4}
\end{subfigure}
\begin{subfigure}{.29\textwidth}
  \centering
  \caption*{~~~~Real~~~~~~~~~~~~~~~~~~~~Generated} \vspace{-6pt}
  \includegraphics[width=.48\linewidth]{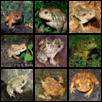}
  \includegraphics[width=.48\linewidth]{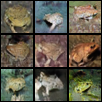}
  \caption{Partition 28}
  \label{fig:apx_cifar_sfig5}
\end{subfigure}
\begin{subfigure}{.29\textwidth}
  \centering
  \caption*{~~~~Real~~~~~~~~~~~~~~~~~~~~Generated} \vspace{-6pt}
  \includegraphics[width=.48\linewidth]{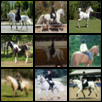}
  \includegraphics[width=.48\linewidth]{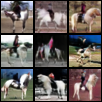}
  \caption{Partition 37}
  \label{fig:apx_cifar_sfig6}
\end{subfigure}
\begin{subfigure}{.29\textwidth}
  \centering
  \caption*{~~~~Real~~~~~~~~~~~~~~~~~~~~Generated} \vspace{-6pt}
  \includegraphics[width=.48\linewidth]{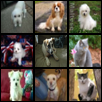}
  \includegraphics[width=.48\linewidth]{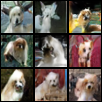}
  \caption{Partition 42}
  \label{fig:apx_cifar_sfig7}
\end{subfigure}
\begin{subfigure}{.29\textwidth}
  \centering
  \caption*{~~~~Real~~~~~~~~~~~~~~~~~~~~Generated} \vspace{-6pt}
  \includegraphics[width=.48\linewidth]{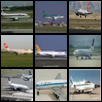}
  \includegraphics[width=.48\linewidth]{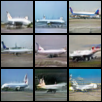}
  \caption{Partition 50}
  \label{fig:apx_cifar_sfig8}
\end{subfigure}
\begin{subfigure}{.29\textwidth}
  \centering
  \caption*{~~~~Real~~~~~~~~~~~~~~~~~~~~Generated} \vspace{-6pt}
  \includegraphics[width=.48\linewidth]{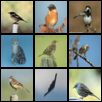}
  \includegraphics[width=.48\linewidth]{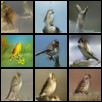}
  \caption{Partition 54}
  \label{fig:apx_cifar_sfig9}
\end{subfigure}
\begin{subfigure}{.29\textwidth}
  \centering
  \caption*{~~~~Real~~~~~~~~~~~~~~~~~~~~Generated} \vspace{-6pt}
  \includegraphics[width=.48\linewidth]{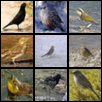}
  \includegraphics[width=.48\linewidth]{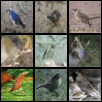}
  \caption{Partition 82}
  \label{fig:apx_cifar_sfig10}
\end{subfigure}
\begin{subfigure}{.29\textwidth}
  \centering
  \caption*{~~~~Real~~~~~~~~~~~~~~~~~~~~Generated} \vspace{-6pt}
  \includegraphics[width=.48\linewidth]{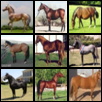}
  \includegraphics[width=.48\linewidth]{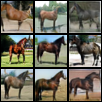}
  \caption{Partition 86}
  \label{fig:apx_cifar_sfig11}
\end{subfigure}
\begin{subfigure}{.29\textwidth}
  \centering
  \caption*{~~~~Real~~~~~~~~~~~~~~~~~~~~Generated} \vspace{-6pt}
  \includegraphics[width=.48\linewidth]{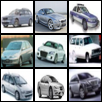}
  \includegraphics[width=.48\linewidth]{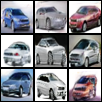}
  \caption{Partition 94}
  \label{fig:apx_cifar_sfig12}
\end{subfigure}
\caption{Extra examples of unsupervised partitioning and their corresponding real/generated samples on CIFAR-10 dataset. \vspace{100pt}}
\label{fig:apx_cifar}
\end{figure*}

\begin{figure*}[b]
\centering
\begin{subfigure}{.35\textwidth}
  \centering
  \caption*{~~~~~~~Real~~~~~~~~~~~~~~~~~~~~~~~~~~Generated} \vspace{-6pt}
  \includegraphics[width=.48\linewidth]{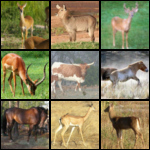}
  \includegraphics[width=.48\linewidth]{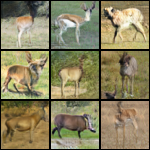}
  \caption{Partition 47}
  \label{fig:apx_stl10_sfig1}
\end{subfigure}
\begin{subfigure}{.35\textwidth}
  \centering
  \caption*{~~~~~~~Real~~~~~~~~~~~~~~~~~~~~~~~~~~Generated} \vspace{-6pt}
  \includegraphics[width=.48\linewidth]{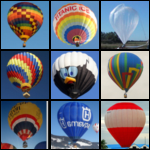}
  \includegraphics[width=.48\linewidth]{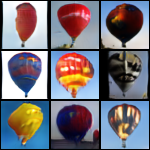}
  \caption{Partition 49}
  \label{fig:apx_stl10_sfig2}
\end{subfigure}
\begin{subfigure}{.35\textwidth}
  \centering
  \caption*{~~~~~~~Real~~~~~~~~~~~~~~~~~~~~~~~~~~Generated} \vspace{-6pt}
  \includegraphics[width=.48\linewidth]{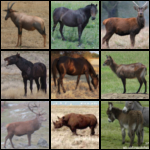}
  \includegraphics[width=.48\linewidth]{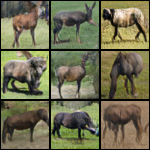}
  \caption{Partition 75}
  \label{fig:apx_stl10_sfig3}
\end{subfigure}
\begin{subfigure}{.35\textwidth}
  \centering
  \caption*{~~~~~~~Real~~~~~~~~~~~~~~~~~~~~~~~~~~Generated} \vspace{-6pt}
  \includegraphics[width=.48\linewidth]{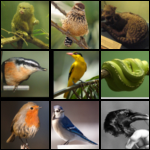}
  \includegraphics[width=.48\linewidth]{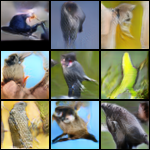}
  \caption{Partition 91}
  \label{fig:apx_stl10_sfig4}
\end{subfigure}
\begin{subfigure}{.35\textwidth}
  \centering
  \caption*{~~~~~~~Real~~~~~~~~~~~~~~~~~~~~~~~~~~Generated} \vspace{-6pt}
  \includegraphics[width=.48\linewidth]{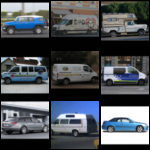}
  \includegraphics[width=.48\linewidth]{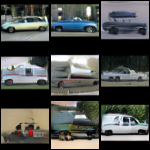}
  \caption{Partition 100}
  \label{fig:apx_stl10_sfig5}
\end{subfigure}
\begin{subfigure}{.35\textwidth}
  \centering
  \caption*{~~~~~~~Real~~~~~~~~~~~~~~~~~~~~~~~~~~Generated} \vspace{-6pt}
  \includegraphics[width=.48\linewidth]{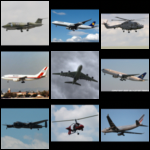}
  \includegraphics[width=.48\linewidth]{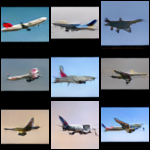}
  \caption{Partition 113}
  \label{fig:apx_stl10_sfig6}
\end{subfigure}
\begin{subfigure}{.35\textwidth}
  \centering
  \caption*{~~~~~~~Real~~~~~~~~~~~~~~~~~~~~~~~~~~Generated} \vspace{-6pt}
  \includegraphics[width=.48\linewidth]{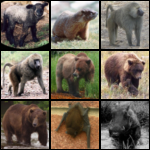}
  \includegraphics[width=.48\linewidth]{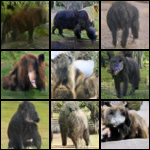}
  \caption{Partition 177}
  \label{fig:apx_stl10_sfig7}
\end{subfigure}
\begin{subfigure}{.35\textwidth}
  \centering
  \caption*{~~~~~~~Real~~~~~~~~~~~~~~~~~~~~~~~~~~Generated} \vspace{-6pt}
  \includegraphics[width=.48\linewidth]{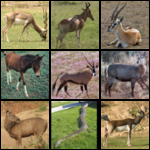}
  \includegraphics[width=.48\linewidth]{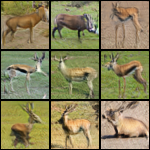}
  \caption{Partition 189}
  \label{fig:apx_stl10_sfig8}
\end{subfigure}
\caption{Extra examples of unsupervised partitioning and their corresponding real/generated samples on STL-10 dataset. \vspace{100pt}}
\label{fig:apx_stl10}
\end{figure*}

We present more samples of our method showing both the partitioner and generative model's performance. Figure~\ref{fig:apx_cifar} and Figure~\ref{fig:apx_stl10} visualize the sample diversity and quality of our method on CIFAR-10 and STL-10.

\subsection{CIFAR-10}

\clearpage

\subsection{STL-10}

\clearpage

\section{Implementation details}
\label{sec:app_implementation_details}

We use two RTX 2080 Ti GPUs for experiments on STL-10, eight V-100 GPUs for ImageNet and a single GPU for all other experiments. 

%\subsection{Architecture}

\textbf{Space partitioner.} For all experiments we use the same architecture for our space partitioner $S$. We use pre-activation Residual-Nets with 20 convolutional bottleneck blocks with 3 convolution layers each and kernel sizes of $3\times3$, $1\times1$, $3\times3$ respectively and the ELU~\cite{clevert2016elu} nonlinearity. The network has 4 down-sampling stages, every 4 blocks where a dimension squeezing operation is used to decrease the spatial resolution. We use 160 channels for all the blocks. We do not use any initial padding due to our theoretical requirements. The negative slope of LeakyReLU is set as 0.2. In fact we can use a soft version of LeakyReLU if it is critical to guarantee the $C^1$ constraint of $\phi$. We train our pretext network for 500 epochs with momentum SGD and a weight decay of 3e-5, learning rate of $0.4$ with cosine scheduling, momentum of $0.9$, and batch size of 400 for CIFAR-10 and 200 for STL-10. The final space partitioner is trained for 100 epochs using Adam~\cite{kingma2014adam} with a learning rate of 1e-4 and batch size of 128. The weights in equation~\ref{eq:cluster} are set to $\alpha = 5$ and $\beta =$~1e-3.

\textbf{Generative model.} Following SN-GANs~\cite{miyato2018spectral} for image generation at resolution 32 or 48, we use the architectures described in Tables~\ref{tab:resnets_cifar10}~and~\ref{tab:resnets_stl10}. Generators/discriminators are different from each other in first-layer/last-layer by having different partition ID embeddings, (which in fact acts as the condition). We use Adam optimizer with a batch size of 100. For the coefficient of guide $\lambda$ we utilized linear annealing during training, decreasing form 6.0 to 0.0001. Both $G$'s and $D$ networks are initialized with a normal $\mathcal{N}(0, 0.02\mI)$. For all GAN's experiments, we use Adam optimizer \cite{kingma2014adam} with $\beta_1 = 0 $, $ \beta_2 = 0.999$ and a constant learning rate 2 for both $G$ and $D$. The number of $D$ steps per $G$ step training is 4. 

For ImageNet experiment, we adopt the full version of BigGAN model architecture~\cite{brock2018large} described in Table~\ref{tab:resnets_imagenet}. In this experiment, we apply the shared class embedding for each CBN layer in G, and feed noise $z$ to multiple layers of $G$ by concatenating with the partition ID embedding vector. Moreover, we add Self-Attention layer with the resolution of 64, and we employ orthogonal initialization for network parameters~\cite{salimans2016improved}. We use batch size of 256 and set the number of gradient accumulations to 8.

\textbf{Evaluation.}
It has been shown that~\cite{chong2020effectively, lucic2018gans} when the sample size is not large enough, both FID and IS are biased, therefor we use N=50,000 samples for computing both IS and FID metrics. We also use the official TensorFlow scripts for computing FID. 

% \textbf{Computationally complexity.} PGMGAN requires additional computation for training the partitioner, and at GANs' training time for computing the guide. On the other hand, due to the higher quality gradients provided by the guide, it requires fewer epoch for the GANs to converge, e.g., over all, compared to conditional model on CIFAR, it takes $\sim$1.6$\times$ in seconds.

% Our implementation can be found in the supplementary file.
% \subsection{Training}
% \textbf{Space partitioner.} We train our pretext network for 500 epochs with momentum SGD and a weight decay of 3e-5, learning rate of $0.4$ with cosine scheduling, momentum of $0.9$, and batch size of 400 CIFAR-10 and 200 for STL-10. The final space partitioner is trained for 100 epochs using Adam~\cite{kingma2014adam} with a learning rate of 1e-4 and batch size of 128. The weights in equation~\ref{eq:cluster} are set to $\alpha = 5$ and $\beta =$~1e-3.

%\textbf{Generative model.} 

% We employ the architectures detailed in table \ref{tab:resnets_cifar10} (for Stacked-MNIST and CIFAR-10) and \ref{tab:resnets_stl10} (for STL-10 dataset). 

% Generators/discriminators are different from each other in first-layer/last-layer by having different partition ID embeddings, (which in fact act as the condition).

\subsection{Additional Experiments:}
Our quantitative results on the (2D-ring, 2D-grid) toy datasets~\cite{lin2018pacgan} are: recovered modes: (8 , 25), high quality samples: (99.8 , 99.8), reverse KL: (0.0006, 0.0034). 

Given the strong performances of recent models (and ours) on these datasets we suffice to these stats. For visualizations of our generated samples related to these two datasets please see Figure~\ref{fig:2d_gan}-left and refer to the appendix of~\cite{liu2020diverse} for other methods.

\textbf{Additional architecture dependent experiment:} Since SelfCondGAN~\cite{liu2020diverse} uses certain features from the discriminator, it is not trivial to adopt to other architectures. Thus, we trained PGMGAN with the same G/D architecture on CIFAR10 yielding an FID of 10.65.

\textbf{Partitioning method}
One way to assess the quality of the space partitioner is by measuring its performance on placing semantically similar images in the same partition. To that end, we use the well-accepted clustering metric Normalized Mutual Information (NMI). NMI is a normalization of the Mutual Information (MI) between the true and inferred labels. This metric is invariant to permutation of the class/partition labels and is always between 0 and 1, with a higher value suggesting a higher quality of partitioning. Table~\ref{tab:clustering_metrics} compares the clustering performance of our method to the-stat-of-the-art partition-based GAN method Liu et. \cite{liu2020diverse}, which clearly shows superiority of our method.

\begin{table}[h]%
  \centering%
  \caption{Comparison of the clustering performance in term of Normalized Mutual Information (NMI), higher is~better.} 
  \vspace{-3mm}
    \resizebox{%
      \ifdim\width>\columnwidth
        \columnwidth
      \else
        \width
      \fi
    }{!}
    {%
    \begin{tabular}{lrrrr}
    \toprule
             & Stacked MNIST & CIFAR-10  &  STL-10 & ImageNet\\
    \midrule
    Self-Cond-GAN~\cite{liu2020diverse}  & 0.3018 & 0.3326  & - & 17.39\\ 
    PGMGAN & 0.4805 & 0.4146 & 0.3911 & 68.57\\ 
    \bottomrule
  \end{tabular}
  } %
  \label{tab:clustering_metrics}
\end{table}

\begin{table}[b]
         \caption{GANs architecture for $32\times 32$ images.}
         \label{tab:resnets_cifar10} 
          \centering
          \small
          \begin{subtable}{.4\textwidth}
              \centering
              {\begin{tabular}{c}
                  \toprule
                  \midrule
                  $z \in \sR^{128} \sim \mathcal{N}(0, I)$ \\
                  Embed(\textit{Partition\textsubscript{ID}}) $\in \sR^{128}$ \\
                  \midrule
                  dense, $ 4 \times 4 \times 256 $ \\
                  \midrule
                  ResBlock up $256$ \\
                  \midrule
                  ResBlock up $256$\\
                  \midrule
                  ResBlock up $256$\\
                  \midrule
                  BN, ReLU, $3\times 3$ Conv, $3$ Tanh \\
                  \midrule
                  \bottomrule
              \end{tabular}}
              \caption{ Generator}
          \end{subtable}
          \begin{subtable}{.4\textwidth}
              \centering
              {\begin{tabular}{c}
                  \toprule
                  \midrule
                  RGB image $x\in \sR^{32 \times 32 \times 3}$ \\
                  \midrule
                  ResBlock down $128$\\
                  \midrule
                  ResBlock down $128$\\
                  \midrule
                  ResBlock $128$\\
                  \midrule
                  ResBlock $128$\\
                  \midrule
                  ReLU, Global sum pooling\\
                  \midrule
                  Embed(\textit{Partition\textsubscript{ID}})$\cdot \vh$ + (linear $\rightarrow$ 1) \\
                  \midrule
                  \bottomrule
              \end{tabular}}
              \caption{\label{tab:dis_resnet_32x32} Discriminator}
          \end{subtable}
\end{table}

\begin{table}[b]
         \caption{GANs architecture for $48\times 48$ images.}
         \label{tab:resnets_stl10} 
          \centering
          \small
          \begin{subtable}{.4\textwidth}
              \centering
              {\begin{tabular}{c}
                  \toprule
                  \midrule
                  $z \in \sR^{128} \sim \mathcal{N}(0, I)$ \\
                  Embed(\textit{Partition\textsubscript{ID}}) $\in \sR^{128}$ \\
                  \midrule
                  dense, $ 3 \times 3 \times 1024 $ \\
                  \midrule
                  ResBlock up $512$ \\
                  \midrule
                  ResBlock up $256$\\
                  \midrule
                  ResBlock up $128$\\
                  \midrule
                  ResBlock up $64$\\
                  \midrule
                  BN, ReLU, $3\times 3$ Conv, $3$ Tanh \\
                  \midrule
                  \bottomrule
              \end{tabular}}
              \caption{\label{tab:gen_resnet_48x48} Generator}
          \end{subtable}
          \begin{subtable}{.4\textwidth}
              \centering
              {\begin{tabular}{c}
                  \toprule
                  \midrule
                  RGB image $x\in \sR^{48 \times 48 \times 3}$ \\
                  \midrule
                  ResBlock down $64$\\
                  \midrule
                  ResBlock down $128$\\
                  \midrule
                  ResBlock down $256$\\
                  \midrule
                  ResBlock down $512$\\
                  \midrule
                  ResBlock $1024$\\
                  \midrule                  
                  ReLU, Global sum pooling\\
                  \midrule
                  Embed(\textit{Partition\textsubscript{ID}})$\cdot \vh$ + (linear $\rightarrow$ 1) \\
                  \midrule
                  \bottomrule
              \end{tabular}}
              \caption{\label{tab:dis_resnet_48x48} Discriminator}
          \end{subtable}
\end{table}

\begin{table}[ht]
         \caption{\label{tab:resnets_imagenet128} GANs architecture for $128\times128$ images. ``$ch$'' represents the channel width multiplier and is set to 96.}
         \label{tab:resnets_imagenet}
          \centering
          \small
          \begin{subtable}{.4\textwidth}
              \centering
              {\begin{tabular}{c}
                  \toprule
                  \midrule
                  $z\in \sR^{120} \sim \mathcal{N}(0, I)$ \\
                  Embed(\textit{Partition\textsubscript{ID}}) $\in \sR^{128}$ \\
                  \midrule
                  Linear $(20+128) \rightarrow 4 \times 4 \times 16 ch $ \\
                  \midrule
                  ResBlock up $16ch \rightarrow 16ch$ \\
                  \midrule
                  ResBlock up $16ch \rightarrow 8ch$\\
                  \midrule
                  ResBlock up $8ch \rightarrow 4ch$\\
                  \midrule
                  ResBlock up $4ch \rightarrow 2ch$\\
                  \midrule
                  Non-Local Block ($64\times 64$)\\
                  \midrule
                  ResBlock up $2ch \rightarrow ch$\\
                  \midrule
                  BN, ReLU, $3\times 3$ Conv $ch\rightarrow 3$ \\
                  \midrule
                  Tanh\\
                  \midrule
                  \bottomrule
              \end{tabular}}
              \caption{\label{tab:gen_resnet_imagenet_128} Generator}
          \end{subtable}
          \begin{subtable}{.4\textwidth}
              \centering
              {\begin{tabular}{c}
                  \toprule
                  \midrule
                  RGB image $x\in \sR^{128 \times 128 \times 3}$ \\
                  \midrule
                  ResBlock down $ch \rightarrow 2ch$\\
                  \midrule
                  Non-Local Block ($64\times 64$) \\
                  \midrule
                  ResBlock down $2ch \rightarrow 4ch$\\
                  \midrule
                  ResBlock down $4ch \rightarrow 8ch$\\
                  \midrule
                  ResBlock down $8ch \rightarrow 16ch$\\
                  \midrule
                  ResBlock down $16ch \rightarrow 16ch$\\
                  \midrule
                  ResBlock $16ch \rightarrow 16ch$\\
                  \midrule
                  ReLU, Global sum pooling\\
                  \midrule
                  Embed(\textit{Partition\textsubscript{ID}})$\cdot \vh$ + (linear $\rightarrow$ 1) \\
                  \midrule
                  \bottomrule
              \end{tabular}}
              \caption{\label{tab:dis_resnet_imagenet_128} Discriminator}
          \end{subtable}
\end{table}

\clearpage

\subsection{ImageNet}

\begin{figure*}[b]
\centering
\begin{subfigure}{.33\linewidth}
  \centering
  \includegraphics[width=0.95\linewidth]{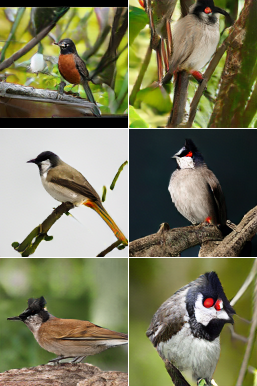}
  \caption{Partition 034}
%   \label{fig:sfig1}
\end{subfigure}%
\begin{subfigure}{.33\linewidth}
  \centering
  \includegraphics[width=0.95\linewidth]{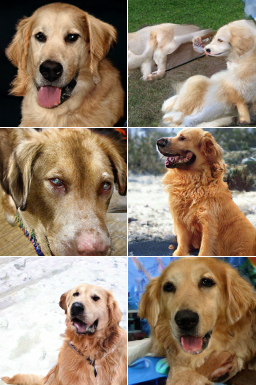}
  \caption{Partition 159}
%   \label{fig:sfig2}
\end{subfigure}
\vspace{8pt}
\begin{subfigure}{.33\linewidth}
  \centering
  \includegraphics[width=0.95\linewidth]{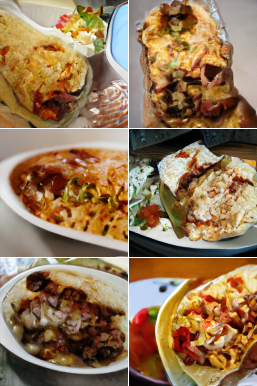}
  \caption{Partition 455}
%   \label{fig:sfig3}
\end{subfigure}%
\begin{subfigure}{.33\linewidth}
  \centering
  \includegraphics[width=0.95\linewidth]{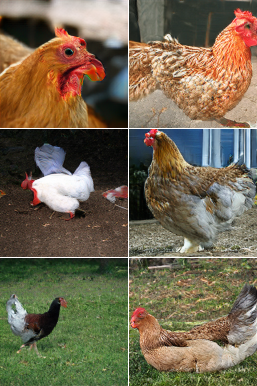}
  \caption{Partition 642}
%   \label{fig:sfig4}
\end{subfigure}%
\vspace{-7pt}
\caption{Examples of generated samples on unsupervised ImageNet~128$\times$128 dataset. \vspace{100pt}}
\vspace{-6pt}
\label{fig:apx_imagenet_I128}
\end{figure*}

\clearpage

\end{appendices}

\end{document}